%% file: main.tex
\pdfoutput=1
\documentclass{article}
\input{config}
\usepackage[accepted]{icml2023}
\usepackage[skip=3pt,belowskip=-6pt,font=small]{subcaption}
\usepackage[skip=3pt,belowskip=-6pt,font=small]{caption}
\makeatletter
\renewcommand{\paragraph}{
  \@startsection{paragraph}{4}
  {\z@}{0ex \@plus 0ex \@minus 0ex}{-1em}
  {\hskip\parindent\normalfont\normalsize\bfseries}
}
\makeatother

\icmltitlerunning{MEWL: Machine word learning}

\begin{document}

\twocolumn[
\icmltitle{\texorpdfstring{\acs{benchmark}}{}: Few-shot multimodal word learning with referential uncertainty}
\icmlsetsymbol{equal}{*}

\begin{icmlauthorlist}
\icmlauthor{Guangyuan Jiang}{pku-ai,pku-yuanpei,bigai}
\icmlauthor{Manjie Xu}{bigai,bit}
\icmlauthor{Shiji Xin}{pku-eecs}
\icmlauthor{Wei Liang}{bit,bit-jiaxing}
\icmlauthor{Yujia Peng}{pku-ai,bigai,pku-psych}
\icmlauthor{Chi Zhang}{bigai}
\icmlauthor{Yixin Zhu}{pku-ai}

\end{icmlauthorlist}

\icmlaffiliation{pku-ai}{Institute for AI, Peking University}
\icmlaffiliation{pku-yuanpei}{Yuanpei College, Peking University}
\icmlaffiliation{pku-eecs}{School of EECS, Peking University}
\icmlaffiliation{pku-psych}{School of Psychological and Cognitive Sciences, Beijing Key Laboratory of Behavior and Mental Health, Peking University}
\icmlaffiliation{bigai}{National Key Laboratory of General Artificial Intelligence, Beijing Institute for General Artificial Intelligence}
\icmlaffiliation{bit}{School of Computer Science \& Technology, Beijing Institute of Technology}
\icmlaffiliation{bit-jiaxing}{Yangtze Delta Region Academy of Beijing Institute of Technology, Jiaxing}

\icmlcorrespondingauthor{Guangyuan Jiang}{jgy@stu.pku.edu.cn}
\icmlcorrespondingauthor{Chi Zhang}{zhangchi@bigai.ai}
\icmlcorrespondingauthor{Yixin Zhu}{yixin.zhu@pku.edu.cn}

\icmlkeywords{word learning, few-shot, concept learning, benchmark, ICML}
\vskip 0.3in
]

\printAffiliationsAndNotice{Code and data: \url{https://github.com/jianggy/MEWL}.}

\setcounter{footnote}{1}

\begin{abstract}
Without explicit feedback, humans can rapidly learn the meaning of words. Children can acquire a new word after just a few passive exposures, a process known as fast mapping. This word learning capability is believed to be the most fundamental building block of multimodal understanding and reasoning. Despite recent advancements in multimodal learning, a systematic and rigorous evaluation is still missing for human-like word learning in machines. 
To fill in this gap, we introduce the \ac{benchmark} benchmark to assess how machines learn word meaning in grounded visual scenes. \ac{benchmark} covers human's core cognitive toolkits in word learning: cross-situational reasoning, bootstrapping, and pragmatic learning.
Specifically, \ac{benchmark} is a few-shot benchmark suite consisting of nine tasks for probing various word learning capabilities. These tasks are carefully designed to be aligned with the children's core abilities in word learning and echo the theories in the developmental literature.
By evaluating multimodal and unimodal agents' performance with a comparative analysis of human performance, we notice a sharp divergence in human and machine word learning. We further discuss these differences between humans and machines and call for human-like few-shot word learning in machines.
\end{abstract}

\section{Introduction}

Learning words and a language is one of the most fundamental stages of human cognitive development, serving as the foundation for other crucial capabilities that come later, such as learning new object categories, forming abstractions of conceptual structures, making generalizations, and developing the ability to communicate \citep{lake2021word,murphy2004big,smith2005development,tenenbaum2011grow}. Remarkably, we acquire the meaning of words rapidly and effortlessly, even without explicit feedback \citep{bloom2001precis}. One striking observation is that young children can understand a novel word's meaning merely from a few examples, also known as fast mapping \citep{carey1978acquiring,heibeck1987word}; a child can learn about 12 words per day by the age of eight \citep{bloom2002children}. These quickly learned words constitute our understanding of the world and the basis of symbol representation for concepts.

\begin{figure}[t!]
    \centering
    \includegraphics[width=\linewidth]{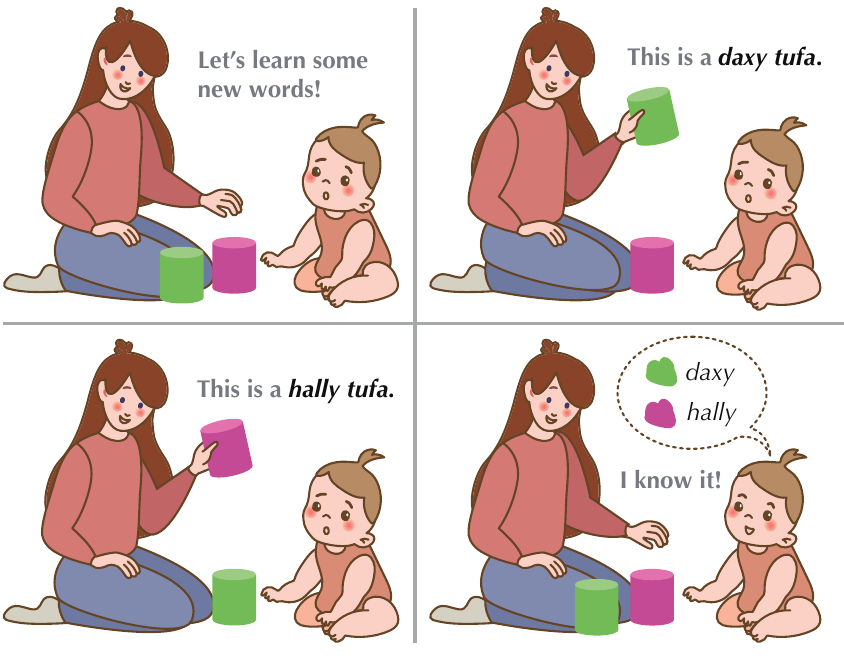}
    \caption{\textbf{Illustration of few-shot word learning.} Children can acquire a novel word after only few exposures using cross-situational information, even with referential uncertainty. In this example, a child induces that \textit{daxy} refers to the color \textcolor{IntroGreen}{green} and \textit{hally} \textcolor{IntroMagenta}{magenta}, all from the experience of a \textit{daxy tufa} (\textcolor{IntroGreen}{green} cylinder) and a \textit{hally tufa} (\textcolor{IntroMagenta}{magenta} cylinder) without explicit guidance.}
    \label{fig:intro}
\end{figure}

Human learning is inherently few-shot and open-ended, even without explicit guidance \citep{landau1988importance,lake2015human}. Children experience substantial referential ambiguity while learning new words, yet they are nevertheless able to comprehend the word-referent mappings; see \cref{fig:intro} for an illustration. How can we learn so many words from so little? Prior developmental studies indicate these capabilities come in many ways.
\begin{itemize}[leftmargin=*,itemsep=0pt,topsep=0pt]
    \item We learn words from co-occurrence in multiple contexts \citep{scott20122}. Children are young statisticians \citep{gopnik1999scientist,abdulla2001babies}; they employ \textbf{cross-situational} statistics \citep{smith2008infants} and perform Bayesian-like inference \citep{tenenbaum1998bayesian} to understand the meaning of words from multiple scenes \citep{xu2007word}.
    
    \item We leverage semantic and syntactic cues to \textbf{bootstrap} novel word learning \citep{quine1960word,pinker2009language}. For instance, we can use familiar relational words to infer an unknown word's meaning: hearing \textit{beef and dax}, we may infer that \textit{dax} is a noun and edible; it may represent a similar kind of food to \textit{beef}.
    
    \item We comprehend word meanings with \textbf{pragmatics}, a social account of word learning with the help of other speakers. The fundamental premise is to leverage informative descriptions of the referent \citep{frank2014inferring,horowitz2016children,stacy2022overloaded}. For example, if we have a \textit{blue cube}, a \textit{blue ball}, and a \textit{green cube} in a line, a speaker would use the word ``\textit{ball}'' to refer to the object in the middle, which is the most informative word to tell them apart \citep{frank2012predicting}.
\end{itemize}

Human-like word learning is quintessential towards building machines that learn and reason like people \citep{lake2017building,zhu2020dark,fan2022artificial}. Despite recent development in language-only and vision-language pre-training, it is still unknown if these models acquire word meaning in a manner similar to that of humans \citep{lake2021word,bender2020climbing,mitchell2023debate}. Concerns have been raised regarding the pre-training paradigm's inability to capture the core components of human language and conceptual structure, such as compositionality \citep{thrush2022winoground}, concept association \citep{yamada2022lemons}, relational understanding \citep{conwell2022testing}, and conceptual meaning \citep{piantasodi2022meaning}. These concerns can be linked to the differences in how humans and machines acquire the primitives of words \citep{fodor1988connectionism,tenenbaum2011grow}. To the best of our knowledge, a systematic and rigorous evaluation for human-like word learning in machines is still missing.

To fill in this gap, we devise the \acf{benchmark} benchmark to assess machine word learning in grounded visual scenes, covering human's core cognitive toolkits in word learning. \ac{benchmark} serves as a testbed for few-shot vision-language reasoning with referential uncertainty. It includes nine tasks covering four types of scenarios: basic attribute naming, relational word learning, number word learning, and pragmatic word learning.

We build \ac{benchmark} in the CLEVR universe \citep{johnson2017clevr}. Each \ac{benchmark} problem consists of six \textit{context images} and corresponding descriptive novel words or phrases (\ie, \textit{utterances}). Agents (either humans or learning algorithms) are tasked to rapidly understand the meaning of novel words from context and choose the option that best matches the target \textit{query image}. These settings closely mimic children's fast cross-situational word learning \citep{goodman2007bayesian,smith2011cross,carey1978acquiring}.

In experiments, we deploy \ac{benchmark} to analyze machines' and humans' ability to perform few-shot word learning under the nine scenarios. We first benchmark machines on \ac{benchmark} by analyzing multimodal (\ie, pre-trained vision-language) and unimodal models (\ie, \acp{llm}). Our experimental results indicate that pre-trained vision-language models struggle to learn word meaning with only a few examples, lagging far behind what humans can do. For \acp{llm}, we turn the word learning problem into a concept binding problem, formulated as in-context learning with images captioned into texts and utterances as labels. \Acp{llm} perform well on attribute and object naming tasks but far worse on all others. Next, we benchmark human performance on \ac{benchmark} for comparison. A comparative analysis reveals misalignment between humans and machines. Finally, we analyze and compare unimodal and multimodal learning algorithms using the rubrics of human-like word learning.

This paper makes three primary contributions:
\begin{enumerate}[leftmargin=*,itemsep=0pt,topsep=0pt]
    \item We highlight the significance of human-like word learning in machines. To support this claim, we devise \ac{benchmark} for probing and comparing few-shot word learning capabilities in machines and humans.
    \item We craft \ac{benchmark} to ensure its similarity to the human counterpart in learning new words. \ac{benchmark} consists of nine tasks, all directly inspired by the established findings in human word learning.
    \item We present a comprehensive benchmark of multimodal and unimodal models on \ac{benchmark}. A comparative analysis of the experimental results shows that large models are generally not human-like in few-shot word learning, calling for future research on building human-like machine models on word and language understanding.
\end{enumerate}

\begin{table*}[tbh!]
    \centering
    \caption{\textbf{Comparison between \ac{benchmark} and prior arts.} We compare \ac{benchmark} and related benchmarks in six dimensions: multimodality, few shot, referential uncertainty, relational reasoning, pragmatic reasoning, and human baseline.}
    \label{tab:comparison}
    \small
    \begin{tabular}{lcccccc}
    \toprule
             & multimodal & few-shot & uncertainty & relation & pragmatic & human             \\
    \midrule
    CLEVR \citep{johnson2017clevr}              & \xmark      & \xmark    & \xmark       & \cmark      & \xmark     & \cmark             \\
    RAVEN \citep{zhang2019raven}                & \xmark      & \cmark    & \xmark       & \cmark      & \xmark     & \cmark             \\
    NLVR \citep{suhr2017corpus}                 & \cmark      & \xmark    & \xmark       & \cmark      & \xmark     & \cmark             \\
    KiloGram \citep{ji2022abstract}             & \cmark      & \xmark    & \xmark       & \xmark      & \xmark     & \cmark             \\
    CURI \citep{vedantam2021curi}               & \cmark      & \cmark    & \cmark       & \cmark      & \xmark     & \xmark             \\
    Fast VQA \citep{tsimpoukelli2021multimodal} & \cmark      & \cmark    & \cmark       & \xmark      & \xmark     & \xmark             \\
    \midrule
    \ac{benchmark} (ours)      & \cmark      & \cmark    & \cmark       & \cmark      & \cmark     & \cmark             \\
    \bottomrule
    \end{tabular}
\end{table*}

\section{Related work}

\paragraph{Word learning in machines}

Despite extensive studies in human word learning, how machines acquire word meaning is almost untouched. Recent attempts use infants' egocentric videos in SAYCam \citep{sullivan2022saycam} and deep learning methods to mimic children's word learning experience \citep{orhan2020self,vong2021cross,rane2022predicting,berger2022computational,vong2020learning,frank2017wordbank,wang2022finding,zhuang2021unsupervised}. While most of these works focus on reverse-engineering the human word learning process, few supporting benchmarks or tasks probe machines' few-shot word learning capabilities. 

Notably, \citet{horst2016novel} introduces the Novel Object and Unusual Name (NOUN) dataset for experimental research. This dataset is relatively small in size (64 images); nonetheless, it supports building word learning algorithms in machines \citep{krishnamohan2020audiovisual,vong2022cross}. In vision-language learning, \citet{tsimpoukelli2021multimodal} introduces Fast VQA, which presents fast concept binding as a new evaluation task for few-shot vision-language models.

In comparison, the proposed \ac{benchmark} benchmark has three distinctions.
\begin{enumerate}[leftmargin=*,itemsep=0pt,topsep=0pt]
    \item \ac{benchmark} focuses not only on basic object categories but also on attribute, relational, numerical, and pragmatic word learning, offering a significantly more comprehensive benchmark suite in human-like word learning.
    \item \ac{benchmark} is akin to human word learning with referential uncertainty, where cross-situational learning is required. This particular setting is almost untouched but is at the core of human-like few-shot word learning. 
    \item Similar to other visual reasoning tasks \citep{johnson2017clevr,barrett2018measuring,depeweg2018solving,edmonds2018human,edmonds2019decomposing,edmonds2020theory,zhang2019raven,zhang2019learning,zhang2021acre,zhang2021abstract,zhang2022learning,nie2020bongard,zhang2020machine,xie2021halma,vedantam2021curi,xu2022est,li2022neural,li2023minimalist}, \ac{benchmark} is light in visual perception but richer in context. It has 37,800 questions, a significantly larger benchmark suite compared to 2,500 in Fast VQA and 64 in NOUN.
\end{enumerate}
In a nutshell, we regard \ac{benchmark} as the first systematic and rigorous benchmark suite for machine word learning.

\paragraph{Human-like few-shot learning}

Children are few-short learners, learning the meaning of new words after merely a single or few exposures \citep{carey1978acquiring,heibeck1987word}.
Modern benchmarks include various few-shot reasoning problems, including the Omniglot challenge \citep{lake2015human,lake2019omniglot}, intelligence measurements \citep{barrett2018measuring,zhang2019raven,chollet2019measure,zhang2020machine}, Bongard problems \citep{depeweg2018solving,nie2020bongard}, causal reasoning \citep{edmonds2018human,zhang2021acre,xu2022est}, and generalization tasks \citep{lake2018generalization,lake2019human,vedantam2021curi,xie2021halma,hsu2022geoclidean,li2022on,li2023minimalist}.

However, these few-shot reasoning problems do not directly tackle the human-like multimodal crux in word learning. Conversely, modern multimodal abstract reasoning benchmarks \citep{kuhnle2017shapeworld,suhr2017corpus,ji2022abstract} are not few-shot by design.
\ac{benchmark} perfectly fills in this gap as a testbed of few-shot multimodal word learning with referential uncertainty. It requires cross-situational grounding of novel words to the learned visual concepts via bootstrapping and pragmatic reasoning. \cref{tab:comparison} provides a comprehensive comparison of \ac{benchmark} with prior arts.

\section{Creating \ac{benchmark}}

\begin{figure*}[t!]
    \centering
    \includegraphics[width=\linewidth]{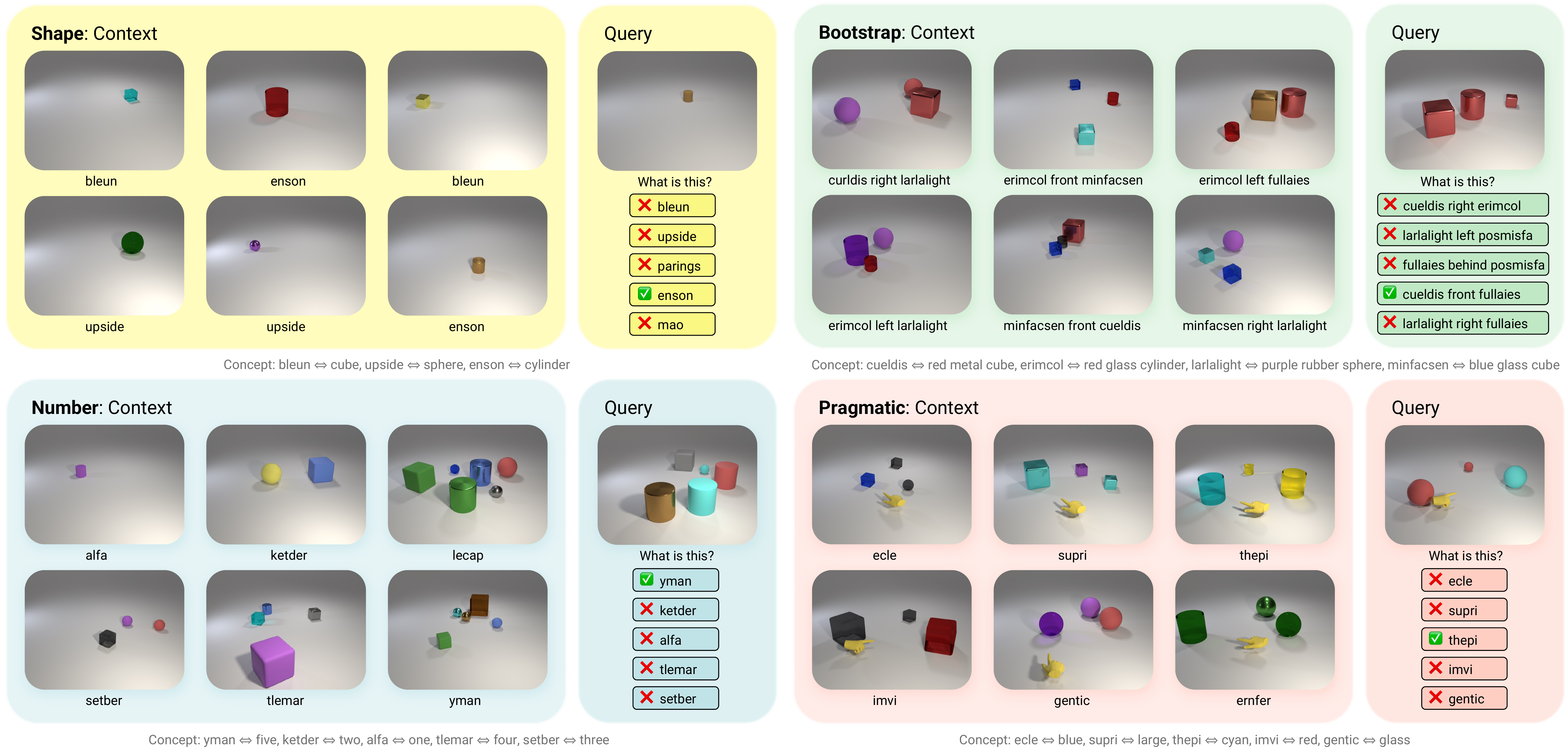}
    \caption{\textbf{Overview of the four categories of tasks in \ac{benchmark}}: (i) basic naming (\eg, \tshape), (ii) bootstrap relational word learning (\eg, \tbootstrap), (iii) learning number words (\eg, \tnumber), and (iv) pragmatic word learning (\ie, \tpragmatic). Each episode consists of six context images and corresponding utterances. Agents need to choose the correct utterance matching the query image out of the given five options based on cross-situational reasoning from the six context panels. Ground-truth word-to-concept mappings are listed below examples (utterance $\leftrightarrow$ concept). Please refer also to \cref{sec:supp:task-examples} for additional examples of the nine tasks.}
    \label{fig:example}
\end{figure*}

When creating \ac{benchmark}, we draw inspiration from and correspondingly highlight these methods in human word learning: cross-situational learning, bootstrapping, and pragmatic word learning. We design nine unique tasks in \ac{benchmark} to comprehensively evaluate alignment between humans and machines: \tshape{}, \tcolor{}, \tmaterial{}, \tobject{}, \tcomposite{}, \trelation{}, \tbootstrap{}, \tnumber{}, and \tpragmatic{}. These tasks cover various aspects:
\begin{itemize}[leftmargin=*,itemsep=0pt,topsep=0pt]
    \item Learn novel words or phrases that represent basic object attributes (\ie, \tshape{}, \tcolor{}, and \tmaterial{}), the objects \textit{per se} (\ie, \tobject{}), and the composition of basic attributes (\ie, \tcomposite{}).
    \item Use familiar words to bootstrap learning novel (spatial) relational words (\ie, \trelation{}) or \textit{vice versa} (\ie, \tbootstrap{}).
    \item Learn counting and number words from one to six (\ie, \tnumber{}).
    \item Use pragmatic cues to learn novel words by assuming the speaker is informative (\ie, \tpragmatic{}).
\end{itemize}

These tasks are crafted to be aligned with the core building blocks in human word learning and echo the theories in the developmental literature \citep{carey1978acquiring,pinker2009language,bloom2002children,scott20122,smith2011cross,horowitz2016children,frank2014inferring}; we detail the setting of each task in \cref{sec:tasks}. As such, \ac{benchmark} constitutes a comprehensive suite for probing how machines learn words' meaning across various few-shot scenarios with referential uncertainty. In MEWL, all nine tasks involve referential uncertainty at varying extents and must be resolved from cross-situational disambiguation. We use the same referential uncertainty concept defined in previous word learning literature: ``For any heard name, there are many candidate referents with variable perceptual properties'' \citep{yu2021infant}.

\ac{benchmark} includes 27,000 problems for training, 5,400 problems for validation, and 5,400 problems for testing.\footnote{In theory, our environment for building \ac{benchmark} allows for the creation of infinite problems. As a result, one can take advantage of this environment and train a foundation model for word learning. However, we contend that this is not how \ac{benchmark} is meant to be used, as the primary objective is to examine the few-shot capability in machine word learning.} These problems are evenly divided among the nine tasks. As shown in \cref{fig:example}, each few-shot problem is an episode consisting of seven images, each containing a few randomly positioned objects. Among them are six context images; each has an utterance consisting of a novel word/phrase describing the image. After seeing context images, a query image is presented with five candidate utterances, with one answer that correctly describes the scene, and therefore formulated as a multiple-choice problem. Following CLEVR \citep{johnson2017clevr}, images are rendered at a resolution of $320 \times 240$ with the Blender Cycles engine \citep{blender2016blender}. Apart from all CLEVR universe objects, we incorporate a glass material for more diversified textures, expanding the space for the \tmaterial{} task. We also include a synthetic yellow rubber hand as the pragmatic pointer for the \tpragmatic{} task. We refer the readers to \cref{sec:supp:dataset,sec:supp:task-examples} for additional details and task examples of \ac{benchmark}. 

To assess the word learning capability in the context of few-shot instead of plain memorization, we create novel words unlikely to be genuine words in the English corpus across episodes. Specifically, we use the 175 most common syllables in the English language to generate more than 5 million pseudo words and associate them with the concepts in the images. Trisyllabic words are generated for \tobject{}, \tcomposite{}, \trelation{}, and \tbootstrap{} tasks, whereas bisyllabic words are generated for \tshape{}, \tcolor{}, \tmaterial{}, \tnumber{}, and \tpragmatic{}. As the words and concepts vary across episodes, the same word can be bound to different concepts in different problems; we assume mutual exclusivity: different novel words have different meanings in each episode \citep{merriman1989mutual}.

\subsection{\ac{benchmark} tasks}\label{sec:tasks}

\paragraph{\tshape{}}

This task tests agents' understanding of novel words for shape concepts from the context panels. \ac{benchmark} has three shapes: \textit{cube}, \textit{sphere}, and \textit{cylinder}. To create word-shape mapping, we randomly assign three unique novel words to the shapes in each episode. When generating context, every context image has one object and a corresponding word as the utterance. Moreover, we control the object's shape to ensure it matches the utterance and leaves other properties (color, size, material) uniformly sampled. For the query panel, we choose one shape for testing. Agents are required to choose the correct word for the shape from the five options that include two distractors.

\paragraph{\tcolor{}}

Similar to the classic fast mapping experiment on children \citep{carey1978acquiring,sandhofer1999learning}, this task studies learning novel words representing colors. \ac{benchmark} has eight distinctive colors: \textit{gray}, \textit{red}, \textit{blue}, \textit{green}, \textit{brown}, \textit{purple}, \textit{cyan}, and \textit{yellow}. We randomly sample three colors out of eight to appear in each episode. Other settings remain the same as in \tshape{}.

\paragraph{\tmaterial{}}

We keep most of the settings unchanged from \tshape{} when designing this task. Instead of naming colors, we name three distinct materials: \textit{rubber}, \textit{metal}, and \textit{glass}.

\paragraph{\tobject{}}

Learning with referential uncertainty is challenging. Inspired by previous studies on infants \citep{smith2008infants,smith2011cross,vong2022cross}, this subset of \ac{benchmark} probes the agents' ability of cross-situational word learning for objects. In this task, novel words bind to the objects \textit{per se} (\ie, the quadruple of size, color, material, and shape). Unlike the \tshape{} task, each image in \tobject{} has three objects and is paired with an utterance consisting of three words. Moreover, an episode has six unique word-object mappings, just within the working memory limit for humans \citep{miller1956magical}. Because there is no one-to-one mapping from the words to objects in an image, agents must perform cross-situational reasoning to determine the correspondence between words and objects.

\paragraph{\tcomposite{}}

This task focuses on learning compositional multi-word phrases and uses syntax to bootstrap the word learning process. In detail, novel words represent basic attributes (\ie, shape, color, and material), and phrases share the same syntax in an episode. The syntax can be any binary combination of the three types of attributes (\eg, the first word represents color, and the second shape). We use the simplified syntax here because syntactic bootstrapping can accelerate the learning of new words \citep{abend2017bootstrapping,gleitman1990structural}. In accordance with the syntax, we selectively name two types of attributes out of three (\eg, color, shape), followed by randomly choosing three instances (\eg, \textit{cyan}, \textit{blue}, \textit{yellow}) for each type of attribute, resulting in a lexicon size $3 + 3 = 6$. To succeed in this task, agents also need to possess systematic generalization because the answer may contain attribute combinations not shown in the context.

\paragraph{\trelation{}}

In this task, we probe agents' capability of learning relational words (\ie, \textit{left}, \textit{right}, \textit{front}, and \textit{behind}). In humans, the understanding of spatial and temporal words is acquired later than object-centric words \citep{friedman1976child}. As temporal words are challenging to evaluate in most models, we only investigate spatial relation words in \ac{benchmark}. To construct spatial relations, we place three objects (with one distractor) in an image and use two familiar English phrases to refer to objects and a novel spatial relation word in between to represent objects' relation (\eg, ``\textit{cyan cube dax red sphere}''). We also replicate the ambiguity when children acquire spatial words. For example, ``\textit{dax}'' can refer to \textit{left} or \textit{front} when inferring from a single image; agents must employ cross-situational reasoning to determine the exact meaning of the spatial words. We design each novel word to appear twice to ensure the problem is solvable. For example, from both \textit{left behind} and \textit{left front}, we understand ``\textit{dax}'' means \textit{left}. Hence, an episode only uses three spatial words, leaving one spatial relation untouched.

\paragraph{\tbootstrap{}}

Recall that we use syntactic cues to bootstrap the learning of attribute words in \tcomposite{}. In this \tbootstrap{} task, we flip the direction by inferring objects' names using familiar relational words. We include all four spatial relation words (\ie, \textit{left}, \textit{right}, \textit{front}, and \textit{behind}) and use novel words to represent objects (similar to the setting in \tobject{}). Each image includes three objects (with one distractor), and the utterance includes relations as cues (\eg, ``\textit{tufa behind dax}''). Agents are tasked to learn the meaning of the six novel words with the help of relational description and choose the correct answer from the five options.

\paragraph{\tnumber{}}

Acquiring numerical words is a giant leap in children's word learning. Instead of acquiring the cardinal principle (the ability to count to infinity, usually acquired at an older stage), we only consider basic learning of counting words \citep{wynn1990children,fuson2012children,piantadosi2012bootstrapping}. This task focuses on how to learn the numerical words, from \textit{one} to \textit{six}. As such, we design the six context images to contain different numbers of objects, ranging from one to six. Each utterance is a unique novel word corresponding to the number of objects in the scene. The query panel includes a random number of objects. To solve the problem, agents need to count how many objects are in the scene and determine the word-number mapping.

\paragraph{\tpragmatic{}}

A critical account in children's word learning is a social-pragmatic theory \citep{tomasello2000social}. Children learn words not only from the cross-situational or linguistic constraints demonstrated in previous tasks but also from inferring communicative intents. In this \tpragmatic{} task, we inspect this pragmatic word learning capability in machines. Inspired by previous studies on human \citep{frank2012predicting,frank2014inferring,fay2010interactive,fay2014iconicity,fay2018create,horowitz2016children,jiang2021individual,jiang2022what,chen2021yourefit,qiu2022emergent}, we design a pragmatic word learning scenario using rendered hands to represent pragmatic pointing. Specifically, every image has a set of three objects and a finger pointing to a referred one, such that the referred object has a unique attribute that can be identified from the context. For example, the targeted object is a \textit{cube} while the other two are \textit{cylinders}. In this case, an informative speaker should use the term ``\textit{cube}'' instead of ``\textit{large cyan metal cube}'' to refer to this object. In \tpragmatic{}, we select six attributes from all available attributes (two sizes, eight colors, three materials, and three shapes) and associate them with unique names. Each of the six context images is paired with a novel attribute word, and we randomly select one attribute to test in the query image. 

For all tasks, we provide the following ground-truth scene information: task name, answer choice, word-to-concept mapping, object types, and coordinates (bounding boxes). \tpragmatic{} is additionally labeled with the pointed object.

\section{Word learning with \ac{benchmark}}

To probe human-like word learning in artificial agents, we examine contemporary models on \ac{benchmark}. Formulating \ac{benchmark} as a few-shot vision-language learning problem, we choose models that fall into two categories: multimodal (vision-language) and unimodal (language-only) models. Please refer to \cref{sec:supp:model} for additional details on model implementation.

\vspace{-0.1in}
\subsection{Multimodal models}

As \ac{benchmark} can be viewed as a vision-language task, we test representative multimodal models: pre-trained vision-language models and models with object-centric embedding.

\paragraph{CLIP}

Contrastive language-image pre-training (CLIP) on large-scale image-caption pairs produces embeddings in a joint image-text embedding space \citep{radford2021learning}, showing superb performance on tasks such as zero-shot classification. We take CLIP's pre-trained vision and text encoder (\ie, CLIP (w/ TE)) to extract features from input images and texts. These features are passed to a Transformer model for classification. We also train a model without using CLIP's pre-trained text encoder (\ie, CLIP (w/o TE)).

\paragraph{Flamingo-1.1B}

Flamingo \citep{alayrac2022flamingo} is designed to tackle few-shot vision-language tasks. It aligns pre-trained vision and language models by training on large-scale multimodal data. Due to its limited availability, we fine-tune an open-sourced 1.1B version, built on the OPT-350M \citep{zhang2022opt} and CLIP (ViT-L/14) \citep{radford2021learning}, pre-trained on the Conceptual Captions (3M) dataset \citep{sharma2018conceptual}.

\paragraph{Aloe}

As all the word learning tasks in \ac{benchmark} are object-based, we additionally test Aloe \citep{ding2020attention}, which uses the Transformer architecture to make predictions based on the learned object embeddings and has demonstrated outstanding performance on previous synthetic visual reasoning tasks \citep{yi2019clevrer,girdhar2019cater,zhang2021acre}. Therefore, we adopt Aloe in \ac{benchmark} with object embeddings learned from MONet \citep{burgess2019monet}.

\vspace{-0.1in}
\subsection{Unimodal models}

\Acp{llm} have been proven to be strong reasoners with few-shot learning abilities. Hence, we test models based on a caption-then-classify paradigm. First, we use a task-specific oracle captioner to parse the input visual scene to generate a scene description. Next, we use language models (\ie, GPT-3.5 \citep{brown2020language} and BERT \citep{devlin2019bert}) to classify the result as a multiple choice problem. Of note, such captions are injected with inductive biases that are precisely needed to solve those tasks, having less uncertainty and ambiguity than images used in the multimodal model. This design drastically simplifies the task difficulty, as it is easier for the unimodal model to map syntactic patterns in the captions to the answer. 
Specifically, inspired by \citet{yang2021AnES}, we prompt GPT-3.5 with a zero-shot multiple-choice template based on full captions generated for the context scenes and the query. We also fine-tune a BERT model on ground-truth captions, resulting in a learned mapping from captions to the answer. Please refer to \cref{sec:supp:caption} for additional details.

\begin{table*}[ht!]
    \caption{\textbf{Performance of baseline models and humans on \ac{benchmark}.}}
    \centering
    \resizebox{\linewidth}{!}{
        \begin{tabular}{lcccccccccc}
            \toprule
                Models & \tshape{} & \tcolor{} & \tmaterial{} & \tobject{} & \tcomposite{} & \trelation{} & \tbootstrap{} & \tnumber{} & \tpragmatic{} & Avg.  \\
                \midrule
                CLIP (w/o TE)     & 16.2 & 18.0 & 19.3 & 17.0 & 22.2 & 20.8 & 18.7 & 19.2 & 20.2 & 19.1 \\
                CLIP (w/ TE)      & 22.0 & 18.8 & 21.0 & 21.2 & 15.0 & 17.8 & 21.0 & 19.5 & 21.5 & 19.8 \\
                Aloe              & 34.2 & 33.2 & 31.0 & 19.5 & 30.5 & 21.5 & 27.5 & 23.3 & 20.8 & 26.8 \\
                Flamingo-1.1B     & 49.3 & 35.3 & 48.5 & 19.2 & 38.2 & 18.8 & 57.3 & 84.2 & 18.0 & 41.0 \\
                \midrule
                BERT              & 94.8 & 98.8 & 97.5 & 19.5 & 97.8 & 22.2 & 62.2 & 21.8 & 99.8 & 68.3 \\
                GPT-3.5             & 96.8 & 82.3 & 87.0 & 98.2 & 88.3 & 20.0 & 45.8 & 22.7 & 26.7 & 63.1 \\
                \midrule
                Human             & 92.4 & 87.2 & 72.7 & 79.1 & 63.5 & 48.7 & 71.0 & 93.9 & 54.8 & 73.7 \\
            \bottomrule
        \end{tabular}
        \label{tab:full_results}
    }
\end{table*}

\begin{figure*}[t!]
    \centering
    \includegraphics[width=.75\linewidth]{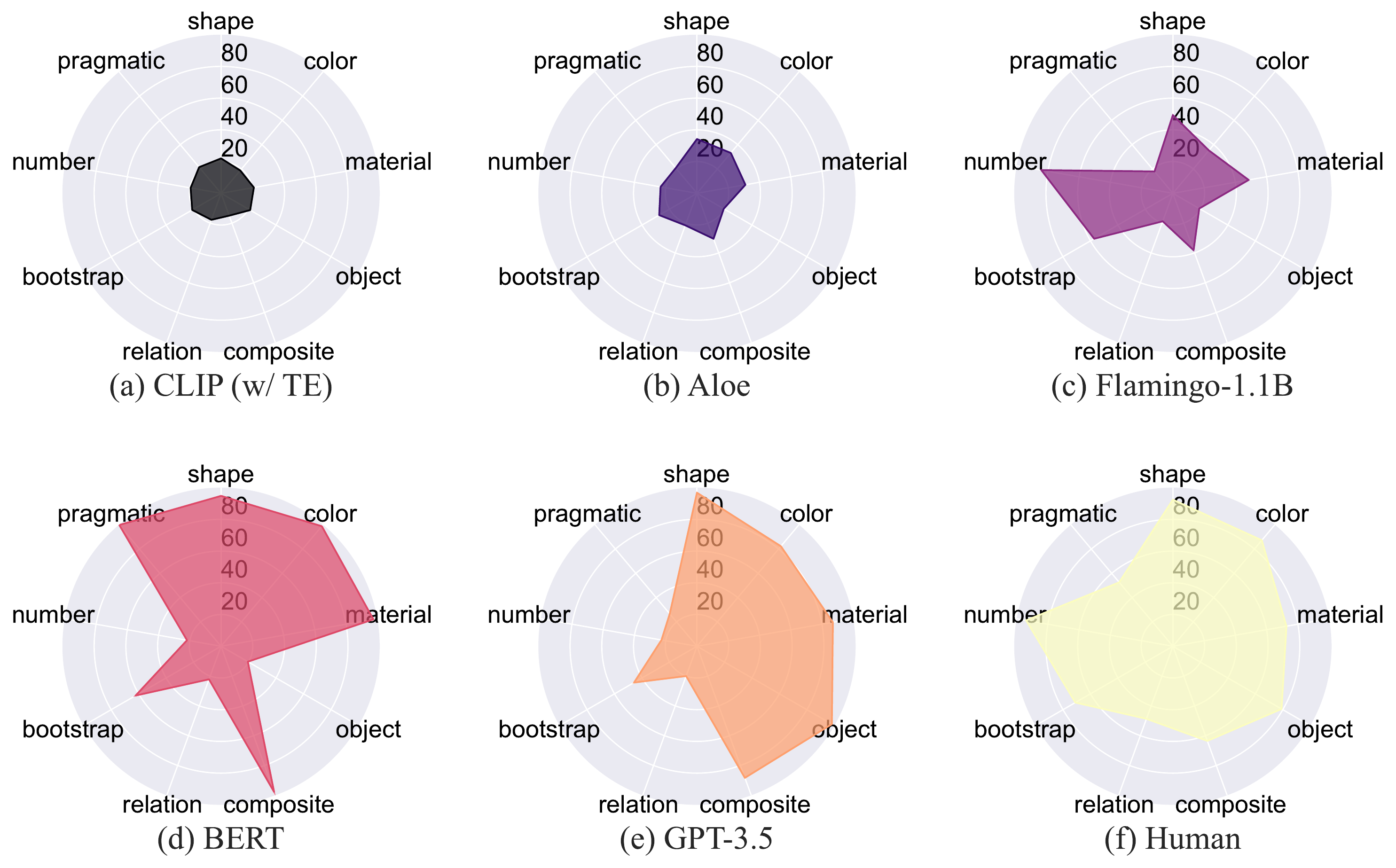}
    \caption{\textbf{Visualization of agents' performance on \ac{benchmark}.}}
    \label{fig:results}
\end{figure*}

\section{Experiments}

We examine few-shot word learning ability by experimenting with machine models and human participants on the proposed \ac{benchmark} benchmark.

\vspace{-0.1in}
\subsection{Experimental setup}

The dataset consists of nine tasks for comprehensively evaluating agents' word learning capabilities. All models except GPT-3.5 are trained on the training sets of all tasks. We report the model performance on the test sets. All experiments run on eight NVIDIA A100 80GB GPUs. GPT-3.5 model is accessed via the OpenAI API (\texttt{text-davinci-003}) with temperature $t=0$.

\vspace{-0.1in}
\subsection{Human study}

To establish a strong baseline to compare with the machines, we looked at how humans performed on the \ac{benchmark} benchmark; this study was approved by the Institutional Review Board (IRB) at Peking University. We designed questionnaires for the human study based on Qualtrics.

Nine questionnaires were constructed, each of which corresponds to a task. To familiarize participants with our study, the Qualtrics workflow first walked them through a step-by-step tutorial. Next, participants were administered tests and attention checks to ensure they comprehended the background and task settings. Specifically, ten questions and two attention check questions are randomly selected from the \ac{benchmark} test set and shuffled in each questionnaire. Attention check questions were designed with a query image identical to one of six context images. Participants who failed these questions or checks were removed.

A total of 271 participants (169 female, mean age 42.8) from the US and UK were recruited from Prolific, an online participant pool, to complete the aforementioned nine tasks and were paid an hourly wage of £6, with a bonus of £0.25-£2. Of the 271 responses collected, 1 failed in familiarization, 52 were removed due to attention check failures, 1 outlier was not counted, and 217 were valid and included in the analysis below. For each of the nine tasks, every participant was presented with a randomly drawn ten-question subset from the task's test set. Please see also \cref{sec:supp:human} for additional details on data processing and significance tests.

\subsection{Results and analysis}

\cref{tab:full_results} summarizes the performance of both machines and humans, with result visualization in \cref{fig:results}.

\paragraph{Multimodal models}

Overall, the best vision-language model is Flamingo-1.1B (41.0\%), only about half as competent as humans (73.2\%). Meanwhile, vanilla transformer models with CLIP features fail catastrophically, achieving only random-level performance on all tasks (less than 20\%). Aloe's object-centric representation helps improve performance to 26.8\% but may fare worse due to limited model capacity and lack of pre-training.

Peeking into task-specific results, we observe that vision-language pre-trained models perform relatively well on basic attribute naming tasks (\ie, \tshape{}, \tcolor{}, \tmaterial{}) but fail to generalize to object relations and reason with pragmatical cues. One interesting observation is that the Flamingo model can solve a small proportion of \tbootstrap{} tasks and some \tnumber{} tasks. This result may be attributed to the Flamingo model being language-model-based, capturing syntactic cues and understanding familiar words to bootstrap word learning.

\paragraph{Unimodal models}

As for unimodal language models, fine-tuned BERT has the best overall performance, with an average performance of 68.3\%. Both BERT and GPT-3.5 achieve outstanding performance on object-level tasks (\ie, \tshape{},\tcolor{}, \tmaterial{},\tobject{}, \tcomposite{}, \tbootstrap{}), yet fail on tasks that require an understanding of more complex relations beyond one-to-one mapping (\ie, \trelation{}, \tnumber{}). Fine-tuned on the training set, the BERT model also performs well on the \tpragmatic{} task, whereas GPT-3.5 (without fine-tuning) fails, indicating that certain capabilities can indeed be learned through task-specific fine-tuning. However, we also want to point out that detailed captions, with strong human bias injected, have been used: We give object-centric captioning to basic attribute naming tasks, relative spatial relations to relational tasks, and the ground-truth pointing to pragmatic tasks. In this sense, the problem is simplified into a translation-like problem, and the challenge of concept abstraction in human word learning is circumvented.

\paragraph{Human performance}

Based on 217 valid responses, our human study suggests that \ac{benchmark} is well-designed and reflects core cognitive skills humans use for word learning. For example, we observe that humans have decent performance on basic naming tasks, with performance ranked \tshape{} $\approx$ \tcolor{} $>$ \tmaterial{} $>$ \tcomposite{}, which echos prior psychological findings of shape bias \citep{landau1988importance} and fast mapping \citep{heibeck1987word}. Humans also perform counting effortlessly. Relational and pragmatic word learning tasks are more challenging than others; relational words often do not have referents to objects, and it is also known to be acquired at the later stage of development \citep{mccune1981cognitive,gentner2005development}. Our human study provides a critical reference for what human-level word learning should demonstrate on \ac{benchmark}.

\section{Discussion}

\subsection{Multimodal \textit{vs.} unimodal}

Comparing multimodal models (\ie, CLIP, Flamingo, and Aloe) and unimodal models (\ie, GPT-3.5 and fine-tuned BERT), we observe that text-based models with ground-truth captioning generally outperform pixel-based ones. This observation in machines seems counter-intuitive as it contrasts with the empirical observations and computational studies on human multimodal learning, which argue that multi-modality boosts the acquisition of words and concepts \citep{clark2006language,smith2005development}. Why and how do contemporary unimodal agents outperform multimodal ones in few-shot word learning? We present some preliminary discussions on this phenomenon in the following.

First, we believe that part of the conceptual role, not all of it, in unimodal language models may be acquired in a way different from humans. Recently, some studies have shown that large language models can encode human-like conceptual structures, even perceptual ones, from unimodal training data \citep{piantasodi2022meaning,abdou2021can}, which are confirmed by experiments on human neural systems \citep{bi2021dual}. In our experiments, GPT-3.5 successfully achieves comparable performance on some basic attribute naming tasks (\ie, \tcolor{}, \tmaterial{}, \tshape{}, \tobject{}, and \tcomposite{}) and yet fails to learn complex relational words (\ie, \tnumber{}, \trelation{}), indicating it already has some conceptual knowledge of shapes, colors, and materials from unimodal training. Nevertheless, GPT-3.5 fails to learn with pragmatic cues, supporting the claim that text-based models cannot infer the communicative word meaning without perceptual grounding \citep{lake2021word}. This leads to the quest for perceptually grounded word learning in machines, to which our \ac{benchmark} contributes.

Second, the unimodal version of \ac{benchmark} is similar to the ``Quine's \textit{Gavagai} Problem'' \citep{quine1960word}. Since we use ground-truth captioners specifically designed for each task, the unimodal language models do not need to undertake the original word learning as humans do with concept induction. Instead, they acquire the meaning of the novel words via few-shot translation from familiar English words, dramatically reducing the difficulty and ambiguity of multimodal word learning. In other words, the unimodal setting is not comparable with the multimodal one. From the experiment of fine-tuning the BERT model, some tasks that do not require complex cross-situational reasoning can be solved with satisfactory performance. By simplifying the problem as unimodal translation, fine-tuning a unimodal model transforms it into a pattern recognition problem, finding hidden statistical patterns from the training data without acquiring actual human-like few-shot word learning capabilities. Hence, we suggest that future work shall not perform specific fine-tuning on the unimodal captioned version of \ac{benchmark} for improving performance but instead use it to compare unimodal and multimodal models. 

\subsection{Humans \textit{vs.} machines}

\paragraph{Efficacy of \ac{benchmark}}

The results of human studies echo previous established developmental studies on human word learning, \eg, the shape bias \citep{landau1988importance} and the shape $\approx$ color $>$ texture (material) relative preference in fast mapping \citep{heibeck1987word}, indicating that our design of \ac{benchmark} indeed captures the essence of human-like word learning.

\paragraph{Failure of learning models}

Experiments with contemporary machine learning models show that they fail to demonstrate human-like word learning capabilities in various tasks. Most multimodal models fail catastrophically, reaching chance-level performance. Some cross-attention vision-language or object-centric models show relatively better performance on specific subtasks. Nonetheless, they still do not match overall human word learning capabilities. Although unimodal large language models achieve outstanding performance on basic naming tasks but fall short in capturing relational and pragmatic word learning. In basic attribute naming tasks, large language models do not show human-like learning (e.g., shape versus texture bias \citep{geirhos2018imagenet,tartaglini2022developmentally}. Crucially, the unimodal paradigm fundamentally differs from human-like multimodal word learning.

\paragraph{Why should machines have human-like word learning capabilities?}

Few-shot word learning is one of the most basic human multimodal reasoning capabilities; it serves as the first step for language acquisition and facilitates learning concepts \citep{clark2006language}. Although recent large-scale vision-language contrastive pre-training \citep{radford2021learning} can be viewed as an approximate form of learning from referential uncertainty, it still diverges much from human-like learning: \eg, the failure in social-pragmatics word learning (\tpragmatic{} task in \ac{benchmark} and \citet{lake2021word}), difficulty in acquiring numerical and relational words (\tnumber{} and \trelation{} tasks in \ac{benchmark}, \citet{radford2021learning}, and \citet{conwell2022testing}), inability to understand compositionality \citep{thrush2022winoground}, and concept association bias \citep{yamada2022lemons}. These problems put it on display that current learning paradigms cannot capture the word meaning in a way similar to humans, leading to an alignment and efficiency problem. Whether human-like word learning should be a path to multimodal AI remains a debate, but it is a fundamental ability for human-AI alignment \citep{yuan2022in}.

Word learning represents a general form of human learning. We learn with referential uncertainty, whereas machines currently do not. We use cross-situational information to support few-shot learning of words and concepts, whereas models currently struggle. We learn with teaching and social-pragmatic cues, whereas artificial agents currently fail to understand. Before bridging the gap, how can we assess the capability of machines to learn words under the same conditions as humans? We take the first step by designing these word learning tasks in machines; \ac{benchmark} is simple and intuitive to support these basic elements in word learning and, in a broader range, human-like learning.

\section{Conclusion}

We propose \acf{benchmark}, a benchmark for human-like few-shot multimodal word learning with referential uncertainty. Inspired by prior developmental studies on how children learn the meaning of words, \ac{benchmark} includes nine carefully designed tasks covering humans' core cognitive toolkits in word learning: cross-situational reasoning, bootstrapping, and pragmatic learning. These tasks make \ac{benchmark} the first comprehensive suite for probing machines' word learning capabilities and echoing human word learning scenarios.

We further examine our tasks on contemporary multimodal and unimodal pre-trained models. By recruiting human participants for comparison on \ac{benchmark}, we found unimodal large language models demonstrate few-shot word learning capabilities on certain subtasks but are still far from human-like. Multimodal vision-language models fail on most tasks, with only the largest language-model-based Flamingo performing better. Together, the results suggest a misalignment of machines' and humans' few-shot word learning capabilities.

We hope \ac{benchmark} serves as the beginning of our journey to building multimodal agents with human-like few-shot learning. Many open problems and opportunities are left for the community to discuss further. For instance, how to build machines that can learn from uncertainty like children do? What role does social-pragmatic learning play in machine learning? Can unimodal \acp{llm} acquire word meaning and conceptual roles in a way similar to humans without perceptual grounding? Will human-like word learning lead to human-like word meaning? As word learning is among the most fundamental cognitive skills for human multimodal understanding, concept learning, and language acquisition, it is undeniably an essential building block for human-like intelligence. We hope our psychologically informed \ac{benchmark} can introduce human-like word learning to machines and motivate future research into this problem.

\paragraph{Broader impact and limitation}

Our \ac{benchmark} launches a new initiative for modern multimodal learning and reasoning: Instead of focusing their performance on pure memorization tasks, we probe their ability of few-shot learning in context, starting with the fundamentals of human multimodal word learning. We hope our work will stimulate future research on developmentally realistic multimodal models that are endowed with the core capabilities and knowledge of human learning. As a first start, we incorporate nine tasks representing four types of word learning into \ac{benchmark}. However, the \ac{benchmark} benchmark is essentially synthetic and devoid of open-vocabulary concepts. As a result, if models are tweaked substantially on the training set, models may find shortcuts, making \ac{benchmark} degenerate into a set of pattern recognition problems. Therefore, we suggest future research on \ac{benchmark} to build core multimodal learning capabilities (inductive biases) in a small-data, developmentally plausible regime.

\paragraph{Acknowledgement}

We would like to thank Yuyang Li (THU) and Nan Jiang (PKU) for their helpful assistance, Liangru Xiang (THU) for constructive discussion, Miss Chen Zhen (BIGAI) for designing the figures, and NVIDIA for their generous support of GPUs and hardware. This work is supported in part by the National Key R\&D Program of China (2022ZD0114900) and the Beijing Nova Program.

{
\balance
\small
\bibliography{reference}
\bibliographystyle{icml2023}
}

\newpage
\appendix
\onecolumn
\renewcommand\thefigure{A\arabic{figure}}
\setcounter{figure}{0}
\renewcommand\thetable{A\arabic{table}}
\setcounter{table}{0}
\renewcommand\theequation{A\arabic{equation}}
\setcounter{equation}{0}
\renewcommand\thealgorithm{A\arabic{algorithm}}
\setcounter{algorithm}{0}
\pagenumbering{arabic}
\renewcommand*{\thepage}{S\arabic{page}}
\setcounter{footnote}{0}

\section{Details for \ac{benchmark} dataset generation}\label{sec:supp:dataset}

This section provides additional details on the generation procedures of \ac{benchmark}.

\subsection{Additional task construction details}

\paragraph{\tshape{}}

There are three shapes in \ac{benchmark}: \textit{cube}, \textit{sphere}, and \textit{cylinder}. We randomly assign three unique novel words to the shapes in each episode to create word-shape mappings. For example, \textit{temmar} is \textit{cylinder}, \textit{subno} is \textit{cube}, and \textit{teis} is \textit{sphere}. When generating the context, we first select a shape (\eg, \textit{cylinder}) and the corresponding word (\eg, \textit{temmar}) as utterance. We then create this context image containing one \textit{cylinder} object and uniformly sample other properties (color, size, material). To make it solvable, we also ensure every panel appears more than once and avoid ambiguous scenarios where two concepts accidentally bind together across all contexts. For the query panel, we choose one shape for testing (with an object of the selected shape as a query image). The candidate options are three novel words (each corresponds to a shape concept) and two dummy words (randomly generated).

\paragraph{\tcolor{}}

There are eight colors in \ac{benchmark}: \textit{gray}, \textit{red}, \textit{blue}, \textit{green}, \textit{brown}, \textit{purple}, \textit{cyan}, and \textit{yellow}. We randomly sample three colors out of eight to appear in each episode. Other settings remain the same as in \tshape{}.

\paragraph{\tmaterial{}}

When designing this task, we keep most of the settings unchanged from \tshape{}. Instead of naming colors, we name three materials: \textit{rubber}, \textit{metal}, and \textit{glass}.

\paragraph{\tobject{}}

In this task, novel words bind to the objects (\ie, the quadruple of size, color, material, and shape). There are 2 (sizes) $\times$ 8 (colors) $\times$ 3 (materials) $\times$ 3 (shapes) = 144 unique objects in \ac{benchmark}. We first sample six unique objects and six novel words to construct images and utterances (\eg, \textit{daythetle} is \textit{mall purple rubber cylinder}, \textit{outsupac} is \textit{small gray metal cube}, \dots, and \textit{peafcol} is \textit{large cyan glass cylinder}). Unlike the \tshape{} task, each image in \tobject{} has three objects. Each utterance has three words representing the three objects in the image (\eg, \textit{daythetle and outsupac and peafcol}). We randomly sample a subset of three objects from the six selected objects to construct a scene. Moreover, we ensure that there are no identical scenes between the six contexts and one query. The five options consist of one ground-truth utterance and four utterances corresponding to object subsets that have not appeared in the context or query.

\paragraph{\tcomposite{}}

This task focuses on learning compositional multi-word phrases and uses syntax to bootstrap the word learning process. In detail, novel words represent attributes (shape, color, and material), and utterance phrases share the same syntax in an episode. In each episode, we first sample two attribute types for naming (\eg, color and shape). We also design an episode-specific syntax (\eg, the first word represents color, and the second word is a shape). For each type of attribute, we randomly choose three instances (\eg, for color, we choose \textit{cyan}, \textit{blue}, \textit{yellow}), resulting in six named attributes (\eg, three colors and three shapes). We map six novel words to these attributes. In each image and the corresponding utterance, we sample one object that satisfies the syntactic constraints (\eg, the object's color must be in the three selected colors, and the object's shape also must be in the three selected shapes, but other attributes of the object are not restricted). We ensure that there are no duplicated attribute pairs (two objects have the same color and shape) among all objects in the images. The rest are the same as \tobject{}.

\paragraph{\trelation{}}

In this task, we probe agents' capability of learning relational words (\ie, \textit{left}, \textit{right}, \textit{front}, and \textit{behind}). To construct spatial relations, we place three objects (with one dummy object) in an image and use two familiar English words to refer to objects and a novel spatial relation word in between to represent objects' relation (\eg, ``\textit{cyan cube dax brown sphere}''). We also replicate the ambiguity when children acquire spatial words. Because spatial locations are ambiguous: For example, \textit{dax} can refer to \textit{left} or \textit{front} when inferring from a single image (the left is often confounded with at least one other orientation). Agents must use cross-situational reasoning to determine the exact meaning of the spatial words. We design each novel word to appear twice to ensure the problem is solvable (\eg, from both \textit{left behind} and \textit{left front}, we can rule out the other confounded orientation and understand \textit{dax} means \textit{left}). Therefore, only three spatial words are used in an episode, leaving one orientation untouched. We use one of the three words (spatial relationships) in the query image.

\paragraph{\tbootstrap{}}

In \tcomposite{}, we have already used some syntactic cues to bootstrap the learning of attribute words. In this \tbootstrap{} task, we take a step further by inferring objects' names using familiar relational words. We include all four spatial relation words (\ie, \textit{left}, \textit{right}, \textit{front}, and \textit{behind}) in familiar English and use novel words to represent objects (similar to the setting in \tobject{}). Six word-concept mappings are required to figure out in every episode. Each image includes three objects (with one dummy object), and the utterance includes relations as cues (\eg, ``\textit{tufa behind dax}''). Agents are tasked to learn the meaning of the six novel words with the help of relational description and choose the correct answer from the five options.

\paragraph{\tnumber{}}

This task focuses on how to learn the number words, from \textit{one} to \textit{six}. In this way, we design the six context images to contain different numbers of objects (ranging from one to six). Each utterance has a unique novel word corresponding to the number of objects in the scene (\eg, \textit{ure} is \textit{one}, \textit{manthe} is \textit{two}, \dots, and \textit{sical} is \textit{six}). The query panel includes a random number (within six) of objects. To solve the problem, agents need to count how many objects are in the scene and figure out the word-number mapping.

\paragraph{\tpragmatic{}}

We design a pragmatic word learning scenario using rendered hands to represent the pointing gesture. In detail, every image has a set of three objects and a finger pointing to a referred object. The referred object has a unique attribute that can be uniquely identified from the context. For example, suppose the targeted object is a \textit{cube}, while the other two are \textit{cylinders}. In that case, an informative speaker should use the term ``\textit{cube}'' instead of ``\textit{large cyan metal cube}'' to refer to this object. In practice, we select six attributes from all available attributes (two sizes, eight colors, three materials, and three shapes) and assign attributes with unique names (\eg, \textit{supcon} is \textit{sphere}, \textit{fuly} is \textit{large}, \dots, and \textit{mainder} is \textit{purple}). Each of the six context images represents a novel attribute word, while we randomly select one attribute for the query image. To generate three objects in a scene, we first sample one base object and modify this base object to become a referred object with a unique attribute. We then modify different attribute types to construct the third object. Agents must understand the correspondence of novel words to referred attributes. The rendering code for referring objects is based on \citet{jiang2020generative}.

\section{Captioning and text input for unimodal models}\label{sec:supp:caption}

For unimodal language models, we use ground truth scene caption for each figure as the input. In this section, we describe the caption generation process and provide some examples of the generated captions.

We use different captions for different types of questions.

\begin{figure}[ht!]
    \centering
    \begin{subfigure}[t]{0.31\linewidth}
        \centering
        \includegraphics[width=\linewidth]{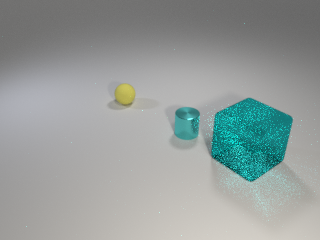}
        \caption{(Object) Caption: A small cyan metal cylinder and a small yellow rubber sphere and a large cyan glass cube.}
        \label{fig:supp:cap_object}
    \end{subfigure}
    \hfill
    \begin{subfigure}[t]{0.31\linewidth}
        \centering
        \includegraphics[width=\linewidth]{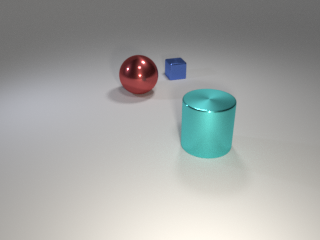}
        \caption{(Spatial) Caption: The large red metal sphere is in front of the small blue metal cube and behind the large cyan metal cylinder. The small blue metal cube is on the left of the large cyan metal cylinder and on the right of the large red metal sphere.}
        \label{fig:supp:cap_relation}
    \end{subfigure}
    \hfill
    \begin{subfigure}[t]{0.31\linewidth}
        \centering
        \includegraphics[width=\linewidth]{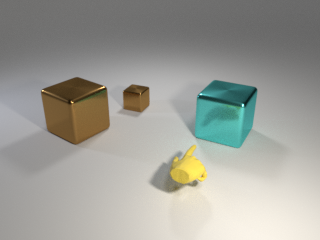}
        \caption{(Pragmatic) Caption: A large brown metal cube and a small brown metal cube and a large cyan metal cube and a small yellow rubber arrow. And a finger is pointing to the large cyan metal cube.}
        \label{fig:supp:cap_pragmatic}
    \end{subfigure}
    \vspace{5pt}
    \caption{\textbf{Examples of generated captions.}}
\end{figure}

The \tobject{}, \tshape{}, \tcolor{}, \tmaterial{}, \tcomposite{}, and \tnumber{} tasks require complete descriptions of all objects in the scene, including their attributes (\ie, color, shape, material, size) to ensure all relevant information is provided. An example of the generated captions is shown in \cref{fig:supp:cap_object}.

The \trelation{} and \tbootstrap{} tasks require knowledge of spatial relations between objects in the scene. To facilitate this, captions for these tasks include the relative position of objects using terms such as ``\textit{front}'', ``\textit{behind}'', ``\textit{left}'', and ``\textit{right}'' along with detailed descriptions of the objects. An example of the generated captions is shown in \cref{fig:supp:cap_relation}.

The \tpragmatic{} task necessitates not only knowledge of the objects and their descriptions but also the identification of the object being pointed at in the scene. An example of the generated captions is shown in \cref{fig:supp:cap_pragmatic}.

To perform the task, we construct the final text input by combining three elements: i) a prompt that specifies the problem (\ie ``\textit{Please name the target object according to the above context.}''). ii) captions for each figure in the context and their associated utterances. iii) captions for the query image along with the provided options.

A full example text input for a problem in \tpragmatic{} can is shown below:

\fbox{\begin{minipage}{\linewidth}
Please name the target object according to the above context.

Context: A small cyan metal cylinder and a small cyan rubber cylinder and a small brown metal cylinder and a small yellow rubber arrow. And a finger is pointing to the small brown metal cylinder.
Name: enre

Context: A large brown metal sphere and a large brown metal cylinder and a large brown rubber sphere and a small yellow rubber arrow. And a finger is pointing to the large brown metal cylinder.
Name: taward

Context: A large brown metal cube and a small brown metal cube and a large cyan metal cube and a small yellow rubber arrow. And a finger is pointing to the large cyan metal cube.
Name: facset

Context: A large brown rubber sphere and a large brown rubber cube and a large brown glass cube and a small yellow rubber arrow. And a finger is pointing to the large brown glass cube.
Name: facov

Context: A small red metal cube and a small purple metal cube and a small red glass cube and a small yellow rubber arrow. And a finger is pointing to the small purple metal cube.
Name: alim

Context: A small green glass sphere and a small green rubber sphere and a large green glass sphere and a small yellow rubber arrow. And a finger is pointing to the large green glass sphere.
Name: tedfac

Context: A small yellow rubber cube and a large yellow rubber cube and a large purple rubber cube and a small yellow rubber arrow. And a finger is pointing to the large purple rubber cube.
Name: \texttt{[Option]}
\end{minipage}
}
\texttt{[Option]} is a candidate utterance (\eg, ``\textit{enre}'', ``\textit{tedfac}'', ``\textit{facset}'', ``\textit{alim}'', or ``\textit{facov}'').

\section{Experimental details}

In this section, we describe the experimental details for the baseline models used in the paper.

For vision-language models, we use the image and the corresponding text as input. While for language-only models, we first use a captioner to parse the image into full scene descriptions, then use the scene descriptions and utterances as input.

\subsection{Model details}\label{sec:supp:model}

\paragraph{CLIP}

The CLIP model utilizes a pre-trained image encoder (ViT-B/16) to extract features from images. Text features are calculated using either the text encoder of CLIP-ViT-B-16 (for CLIP (w/ TE)) or CLIP's token embedding (for CLIP (w/o TE)). The resulting features are concatenated in the format of \texttt{[image1, utterance1, image2, utterance2, \ldots, image6, utterance6, image query, option1, option2, \ldots, option5]}. These input features are then passed through a 6-layer Transformer model with an MLP head for classification. 
We freeze the CLIP when training. The model is trained on the training set for $600$ epochs, dropout $0.1$, batch size $64$, learning rate $1\times10^{-4}$, and AdamW optimizer (weight\_decay $0.01$).

\paragraph{Aloe}

The Aloe model employs the MONet architecture as provided by \citet{engelcke2020genesis}. The MONet model is pre-trained on the training set images resized into $128\times 128$ for $600$ epochs with Adam ($\text{lr}=1\times10^{-5}$), $7$ slots, and a latent dimension of 16. We take the mean as the object feature of the figure and use this feature as the visual input for the Transformer. As for the text input, we embed each word using a trainable embedding as follows: i) If the utterance is a typical English word (e.g. \textit{yellow}, \textit{metal}, \textit{and}), the utterance is directly encoded by the embedder. ii) If the word is a novel one, it is embedded as a random placeholder. The visual inputs are concatenated with text embeddings of the context utterances and choices in a similar format as the CLIP model inputs. These inputs are then passed through a Transformer model with a head size of $8$, a latent dimension of $512$, and trained for $600$ epochs using the Adam optimizer with a learning rate of $1\times10^{-4}$ for classification. Hyperparameters for the Transformer model are as follows: training $200$ epochs, learning rate $5\times10^{-5}$, Adam optimizer ($\beta_1 = 0.9, \beta_2 = 0.999, \epsilon = 1\times10^{-8}$, linear learning rate decay, and batch size $128$.

\paragraph{Flamingo-1.1B}

Since the Flamingo models' pre-trained weights are not accessible, we use open-sourced implementations of the Flamingo model\footnote{\url{https://github.com/lucidrains/flamingo-pytorch} and \url{https://github.com/dhansmair/flamingo-mini}}. Specifically, we use a 1.1B Flamingo model built upon OPT-350M and CLIP pre-trained ViT-L/14 model. The model is pre-trained on the Conceptual Captions dataset. 

For \ac{benchmark} tasks, we formulated it as a multiple-choice problem (similar to how GPT performs multiple-choice tasks). Specifically, we add a binary classifier head to Flamingo's last layer outputs. Meanwhile, we concatenate episodic interleaved image and utterance pairs, the query image, and one option (candidate utterance) as input. We feed each episode five times to get the logits corresponding to the five options and pass through a softmax layer to get the final answer. We use cross-entropy loss for training. Hyperparameters are as follows: training steps $30000$ ($\approx 106$ epochs), learning rate $5\times10^{-5}$, Adam optimizer ($\beta_1 = 0.9, \beta_2 = 0.999, \epsilon = 1\times10^{-8}$, linear learning rate decay, and batch size $96$.

\paragraph{GPT-3.5}

We use the \texttt{text-davinci-003} model provided by the OpenAI API. Formulated as a few-shot multiple choice question answering problem, we concatenate all the image captions and utterances as inputs. Formally following four steps: i). We caption six context images and a query image into \texttt{caption1, caption2, caption3, \dots, caption6, query\_caption}. ii). We concatenate them with the corresponding utterances (context) to get the context input $\mathcal{C}=$\texttt{[caption1, utterance1, caption2, utterance2, \dots, caption6, utterance6]}. iii). We then construct five inputs for GPT-3.5 by concatenating context input, query caption, and a possible option (\eg, \texttt{[$\mathcal{C}$, query\_caption, option1]}, \texttt{[$\mathcal{C}$, query\_caption, option2]}, \dots, \texttt{[$\mathcal{C}$, query\_caption, option5]}). iv). Finally, we feed the prompt to GPT-3.5 and choose the one with the largest log probability as the answer.

\paragraph{BERT}

For the BERT model, we follow standard practice for utilizing BERT as a multiple-choice question answering; see Section 4.4 of \citet{devlin2019bert} for details. We first generate ground truth captions for each figure using the captioner and construct the question input \texttt{[caption1, utterance1, caption2, utterance2, \dots, caption6, utterance6, query\_caption]}. Then, we construct five input sentences by concatenating the question input and a candidate option. We adopt a linear scoring head and a softmax layer on the last layer's \texttt{[CLS]} hidden state to calculate the class probability.

We fine-tune a BERT-base model on the training set for 200 epochs, with learning rate $5\times10^{-5}$, Adam optimizer ($\beta_1 = 0.9, \beta_2 = 0.999, \epsilon = 1\times10^{-8}$, linear learning rate decay, and batch size $64$.

\section{Human study}\label{sec:supp:human}

271 participants were recruited from Prolific (169 female; mean age 42.8) for the nine tasks. All of the participants are from UK or USA and have a Bachelor's degree or higher. The participants were paid an hourly wage of £6 (with a bonus of £0.25-£2). This study has been approved by an IRB. 270 of 271 responses are accepted (one failed in familiarization),  217 of which are valid (52 removed due to failures in attention checks, and one removed due to outlier).

\subsection{Data processing}

Responses from participants who failed attention check questions were removed when measuring human performance. Besides, Grubbs' tests were performed with the significance level $\alpha = 0.5$ in each group to remove outlier results. Only one outlier was detected and removed.

\subsection{Tests of statistical significance}

We used a \textit{t}-test to determine if one task is significantly easier than the other. The \textit{t}-test was performed between any two groups of human results with the significance level $p < .05$ (one-tailed). Our null hypothesis is that the group of results with a higher mean is not significantly better than the other group. the The average of human performance and \textit{t}-test results are shown in \cref{tab:human_results} and \cref{tab:supp:t-test}, respectively.

\begin{table}[ht!]
    \centering
    \caption{\textbf{Performance of humans on \ac{benchmark}.}}
    \label{tab:human_results}
    \small
    \resizebox{\linewidth}{!}{
        \begin{tabular}{lcccccccccc}
            \toprule
                ~ & \tshape{} & \tcolor{} & \tmaterial{} & \tobject{} & \tcomposite{} & \trelation{} & \tbootstrap{} & \tnumber{} & \tpragmatic{} & Avg.  \\
                \midrule
                Human             & 92.4 & 87.2 & 72.7 & 79.1 & 63.5 & 48.7 & 71.0 & 93.9 & 54.8 & 73.7 \\
            \bottomrule
        \end{tabular}
    }
\end{table}

\begin{table}[ht!]
    \centering
    \caption{\textbf{The \textit{p}-values of human results on the \ac{benchmark}}. \textcolor{Blue}{Blue} indicates that the task with a higher average is significantly easier for humans than the other, while \textcolor{Red}{Red} indicates that there is no significant difference in difficulty between the two tasks.}
    \label{tab:supp:t-test}
    \small
    \begin{tabular}{ccccccccc} 
        \toprule
                            & \tcolor{}            & \tmaterial{}         & \tobject{}            & \tcomposite{}      & \trelation          & \tbootstrap{}             & \tnumber{}           & \tpragmatic{}             \\ 
        \midrule
        \tshape{}    & \textcolor{red}{.098} & \textcolor{blue}{$<$ .001} & \textcolor{blue}{.003} & \textcolor{blue}{$<$ .001} & \textcolor{blue}{$<$ .001} & \textcolor{blue}{.001} & \textcolor{red}{.305}  & \textcolor{blue}{$<$ .001}  \\ 
        \tcolor{}     &           -            & \textcolor{blue}{.006} & \textcolor{red}{.065}  & \textcolor{blue}{$<$ .001} & \textcolor{blue}{$<$ .001} & \textcolor{blue}{.002} & \textcolor{red}{.058}  & \textcolor{blue}{$<$ .001}  \\ 
        \tmaterial{}   &           -            &           -            & \textcolor{red}{.133}  & \textcolor{red}{.082}  & \textcolor{blue}{$<$ .001} & \textcolor{red}{.385}  & \textcolor{blue}{$<$ .001} & \textcolor{blue}{.015}  \\ 
        \tobject{}   &           -            &           -            &           -            & \textcolor{blue}{.009} & \textcolor{blue}{$<$ .001} & \textcolor{red}{.074}  & \textcolor{blue}{.002} & \textcolor{blue}{.002}  \\ 
        \tcomposite{} &           -            &           -            &           -            &           -            & \textcolor{blue}{.016} & \textcolor{red}{.120}  & \textcolor{blue}{$<$ .001} & \textcolor{red}{.150}   \\ 
        \trelation{}  &           -            &           -            &           -            &           -            &           -            & \textcolor{blue}{$<$ .001} & \textcolor{blue}{$<$ .001} & \textcolor{red}{.225}   \\ 
        \tbootstrap{}        &           -            &           -            &           -            &           -            &           -            &           -            & \textcolor{blue}{$<$ .001} & \textcolor{blue}{.021}  \\ 
        \tnumber{}    &           -            &           -            &           -            &           -            &           -            &           -            &           -            & \textcolor{blue}{$<$ .001}  \\
        \bottomrule
    \end{tabular}
\end{table}

\section{More task examples}\label{sec:supp:task-examples}
\input{demo}

\end{document}

%% file: config.tex
\usepackage{microtype}
\usepackage{graphicx}
\usepackage{amsmath,amssymb,mathbbol}
\usepackage{acronym}
\usepackage{enumitem}
\usepackage[pagebackref,breaklinks,colorlinks]{hyperref}
\usepackage{balance}
\usepackage{xspace}
\usepackage{setspace}
\usepackage[dvipsnames,svgnames,x11names]{xcolor}
\usepackage[capitalise,noabbrev,nameinlink]{cleveref}
\usepackage{booktabs,tabularx,colortbl,multirow,multicol,array,makecell}
\usepackage{pifont}
\usepackage{textcomp,scalerel}

\makeatletter
\DeclareRobustCommand\onedot{\futurelet\@let@token\@onedot}
\def\@onedot{\ifx\@let@token.\else.\null\fi\xspace}
\def\eg{\emph{e.g}\onedot} 

\def\ie{\emph{i.e}\onedot}

\makeatother

\input{math_commands.tex}

\graphicspath{{figures/}}

\definecolor{gblue}{HTML}{4285F4}
\definecolor{gred}{HTML}{DB4437}
\definecolor{ggreen}{HTML}{0F9D58}

\def\emojicat{\scalerel*{\includegraphics{cat}}{\textrm{\textbigcircle}}}

\acrodef{llm}[LLM]{Large Language Model}
\acrodef{benchmark}[\protect\emojicat{}\,MEWL]{\underline{M}achin\underline{E} \underline{W}ord \underline{L}earning}

\newcommand{\tshape}{\texttt{shape}}
\newcommand{\tcolor}{\texttt{color}}
\newcommand{\tmaterial}{\texttt{material}}
\newcommand{\tobject}{\texttt{object}}
\newcommand{\tcomposite}{\texttt{composite}}
\newcommand{\trelation}{\texttt{relation}}
\newcommand{\tbootstrap}{\texttt{bootstrap}}
\newcommand{\tpragmatic}{\texttt{pragmatic}}
\newcommand{\tnumber}{\texttt{number}}

\definecolor{MyPurple}{RGB}{111,0,255}
\definecolor{MyBlue}{RGB}{0,111,255}
\definecolor{MyRed}{RGB}{255,0,111}
\definecolor{MyGreen}{RGB}{6,128,67}
\definecolor{MyRdd}{RGB}{255,111,111}
\definecolor{IntroGreen}{RGB}{101,175,85}
\definecolor{IntroMagenta}{RGB}{188,75,137}

\newcommand{\cmark}{\ding{51}}%
\newcommand{\xmark}{\ding{55}}%

%% file: math_commands.tex
\usepackage{amsmath,amsfonts,bm}

\def\eqref#1{equation~\ref{#1}}

\def\1{\bm{1}}

\DeclareMathAlphabet{\mathsfit}{\encodingdefault}{\sfdefault}{m}{sl}
\SetMathAlphabet{\mathsfit}{bold}{\encodingdefault}{\sfdefault}{bx}{n}



%% file: demo.tex
\subsection{\tshape{}}
\begin{figure}[ht!]
    \centering
    \includegraphics[width=0.83\linewidth]{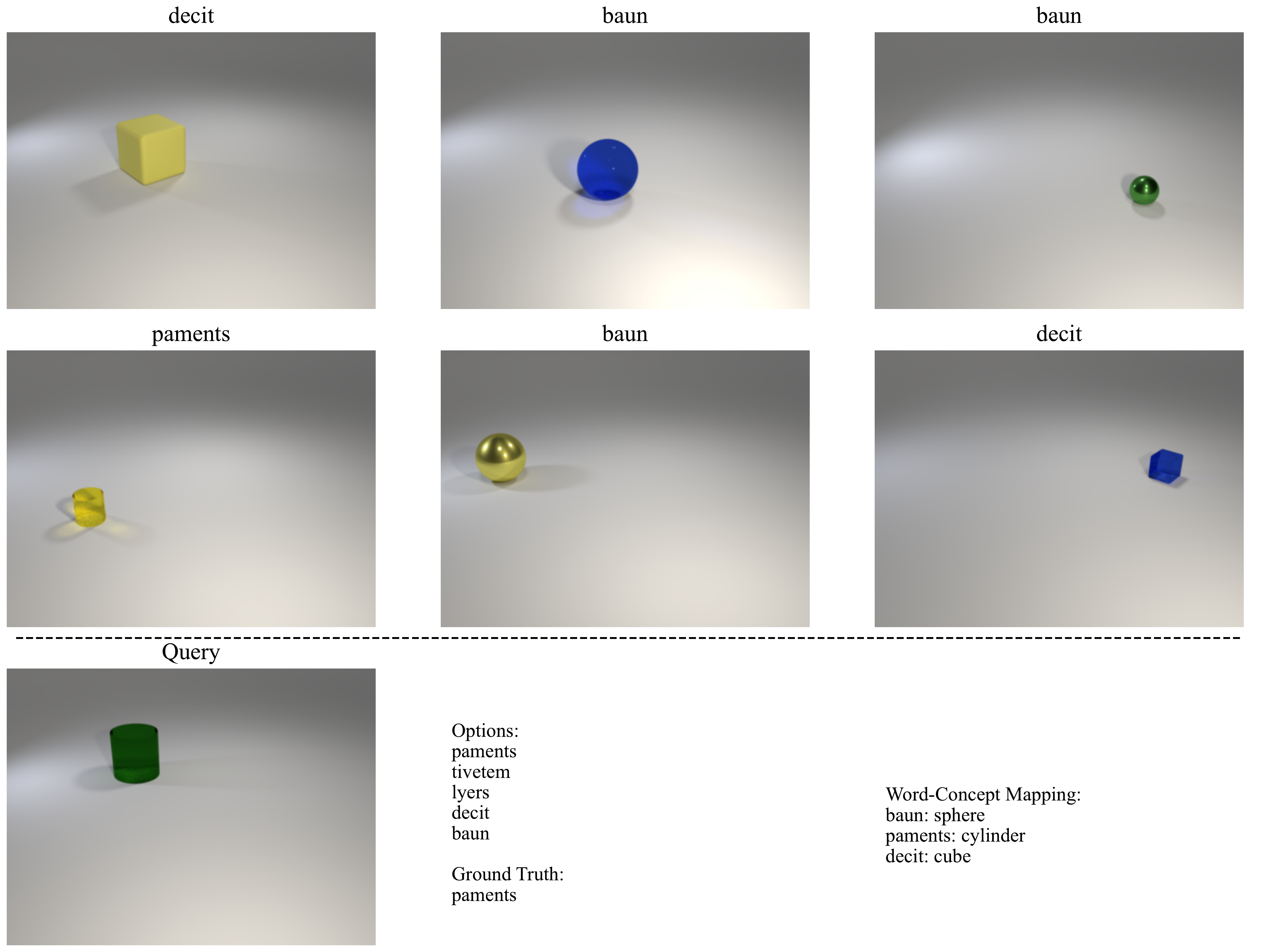}
    \includegraphics[width=0.83\linewidth]{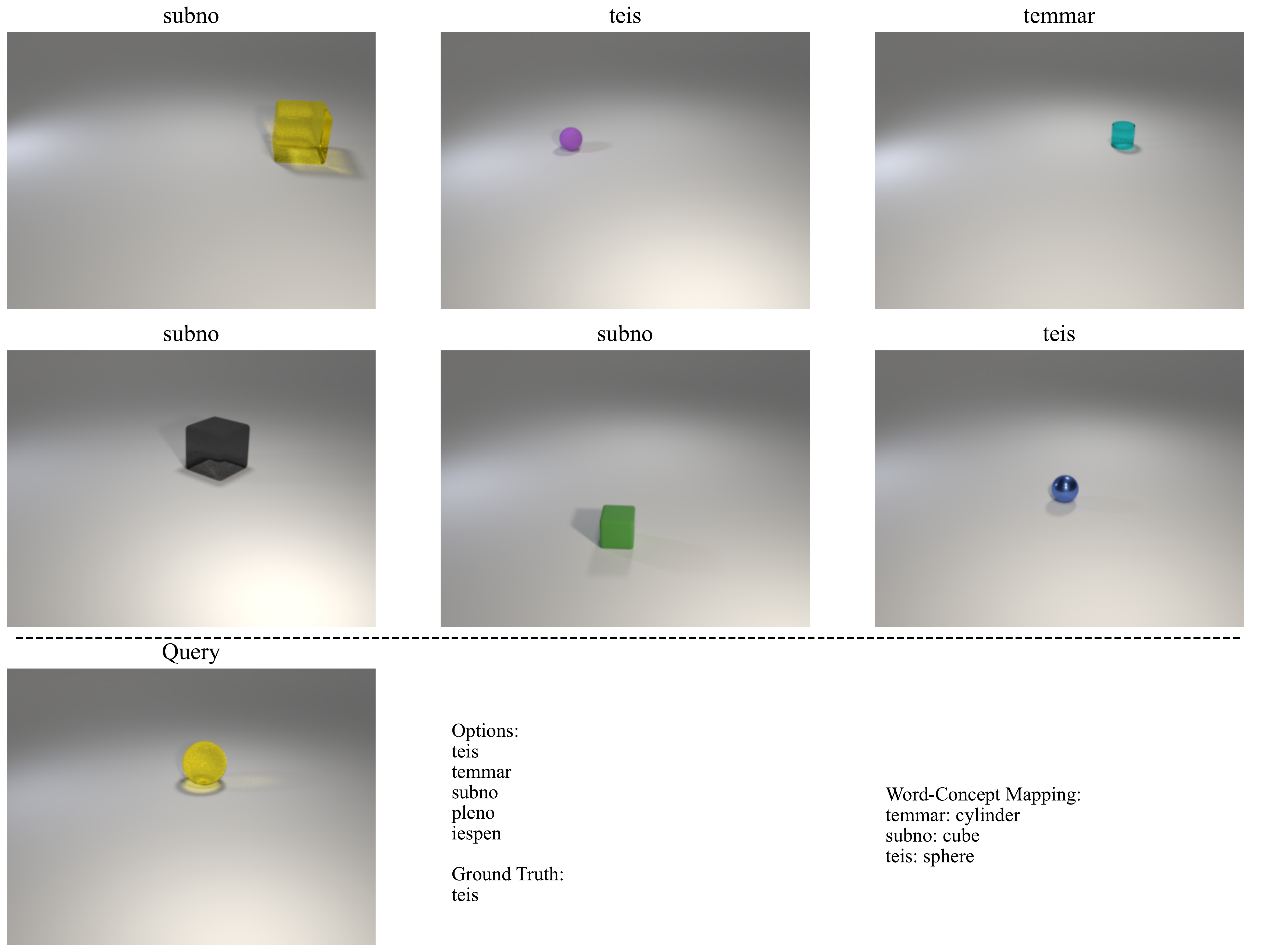}
\end{figure}

\subsection{\tcolor{}}
\begin{figure}[ht!]
    \centering
    \includegraphics[width=0.83\linewidth]{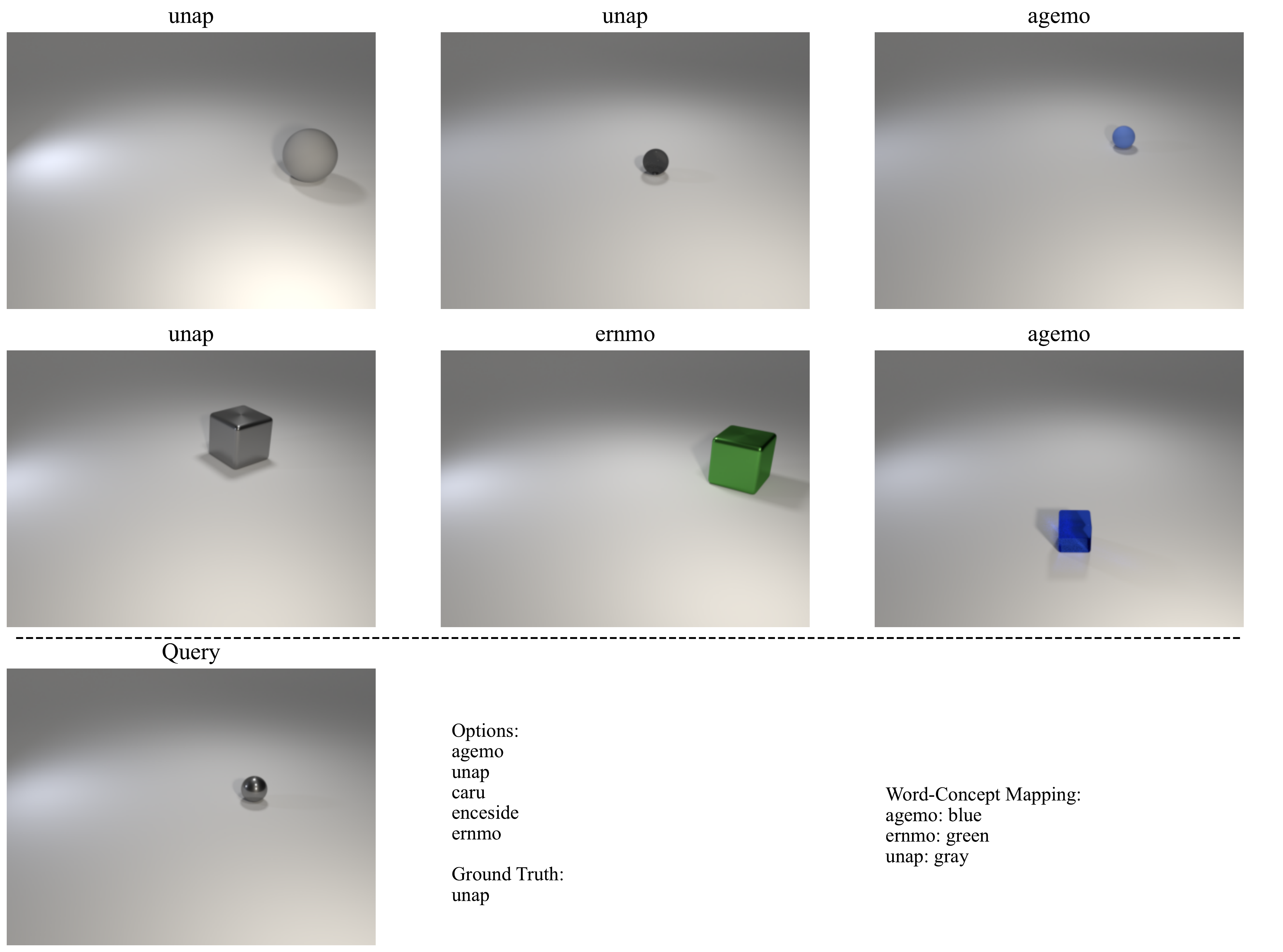}
    \includegraphics[width=0.83\linewidth]{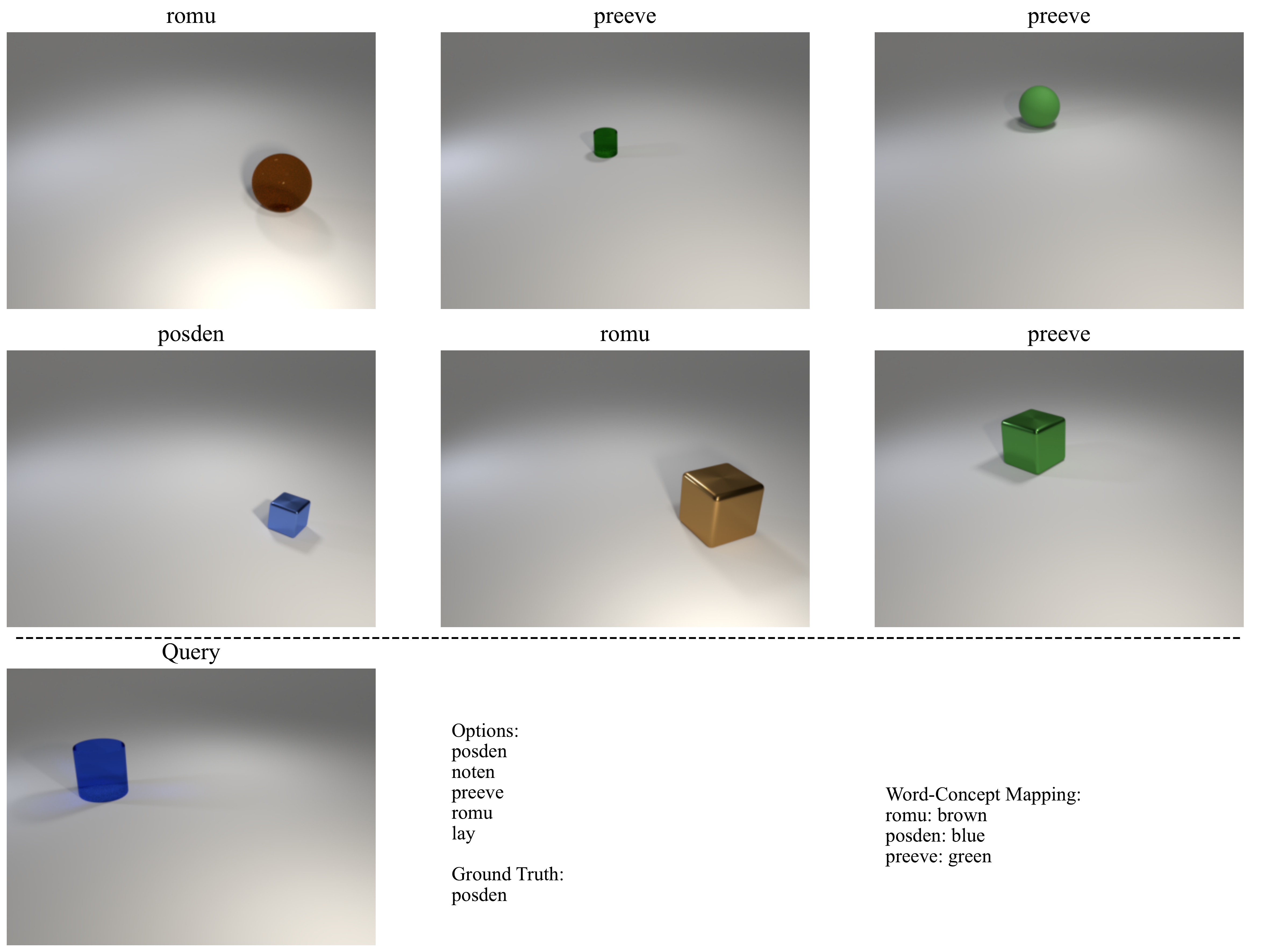}
\end{figure}

\subsection{\tmaterial{}}
\begin{figure}[ht!]
    \centering
    \includegraphics[width=0.83\linewidth]{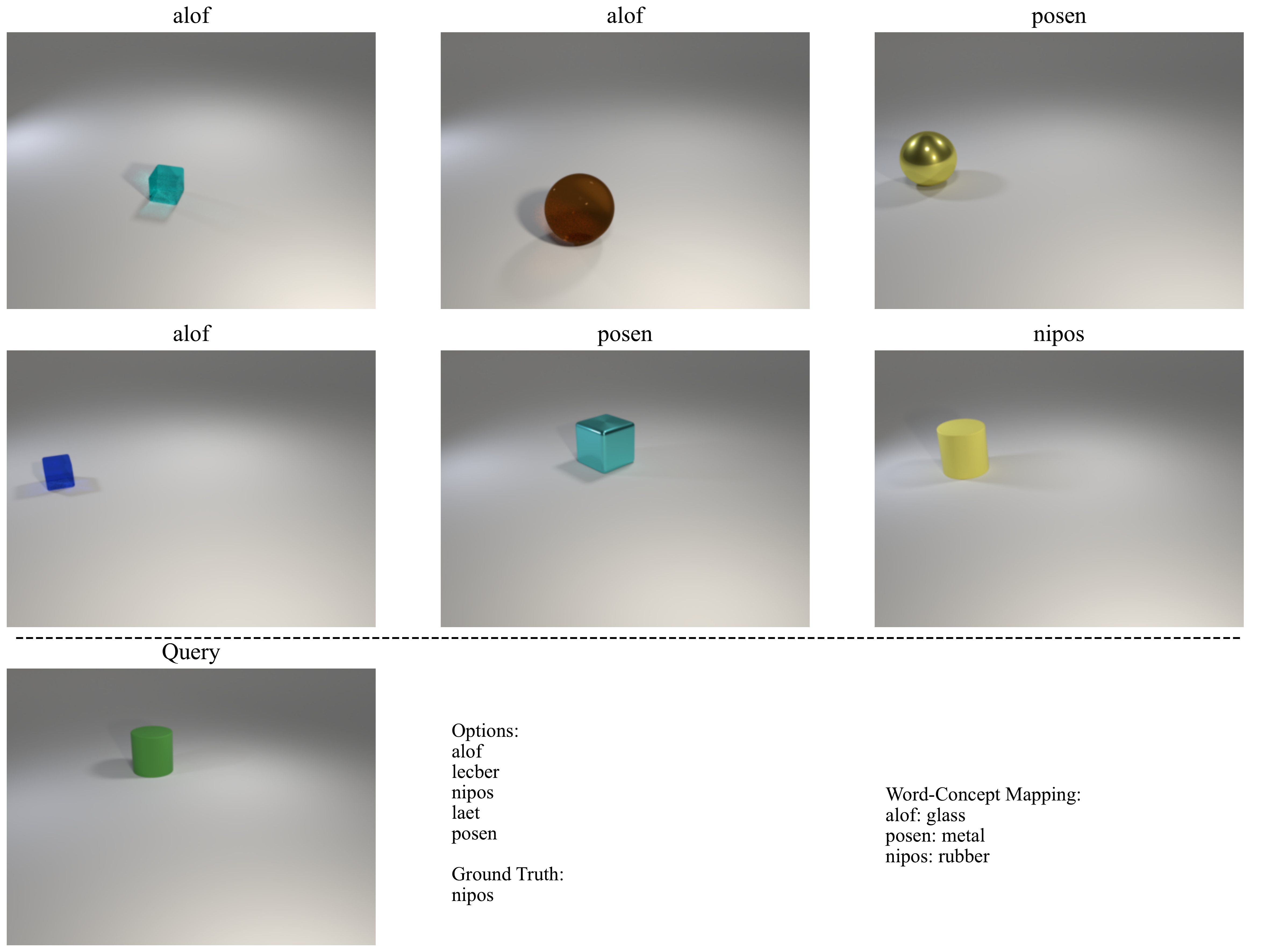}
    \includegraphics[width=0.83\linewidth]{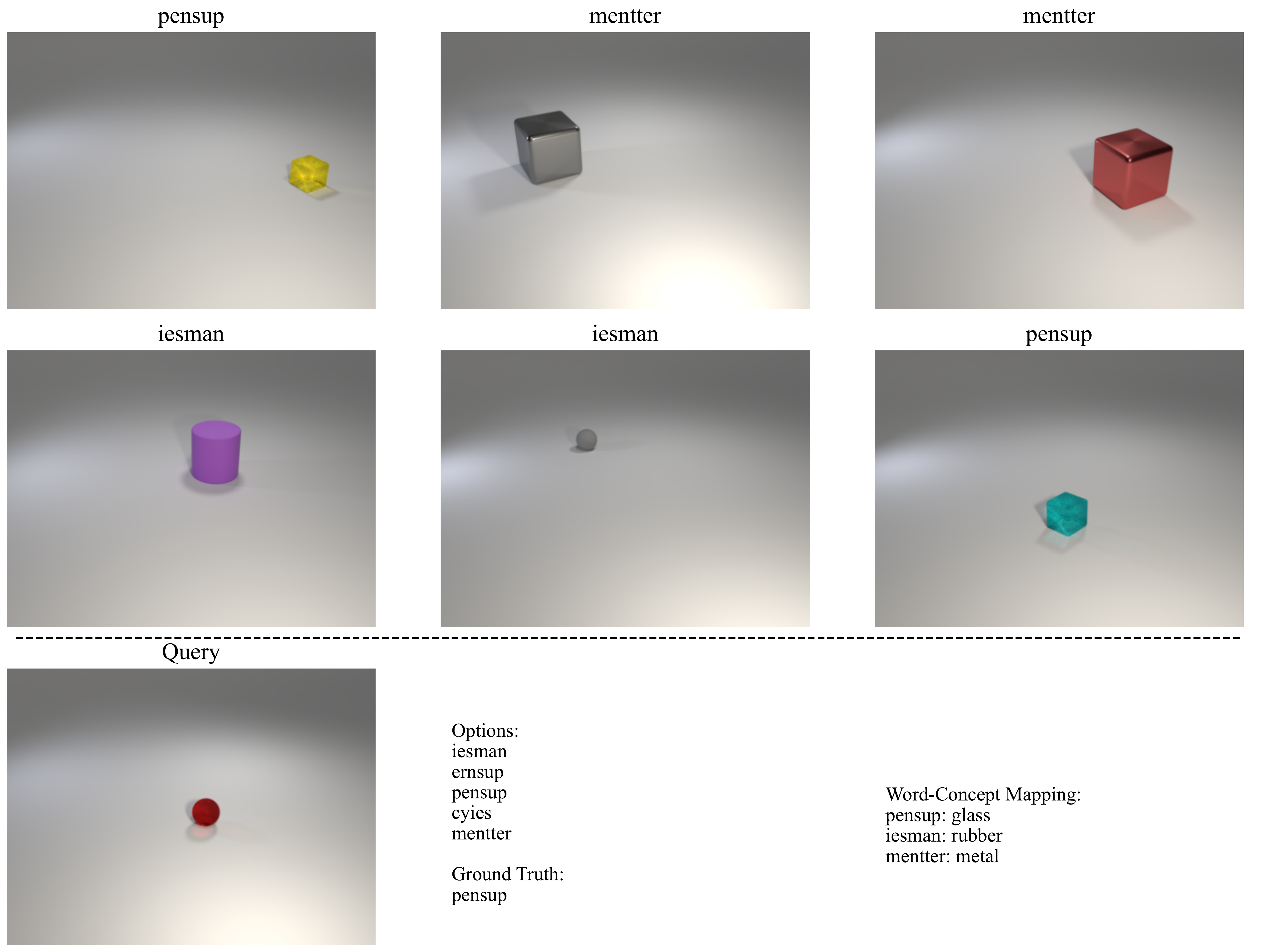}
\end{figure}

\subsection{\tobject{}}
\begin{figure}[ht!]
    \centering
    \includegraphics[width=0.83\linewidth]{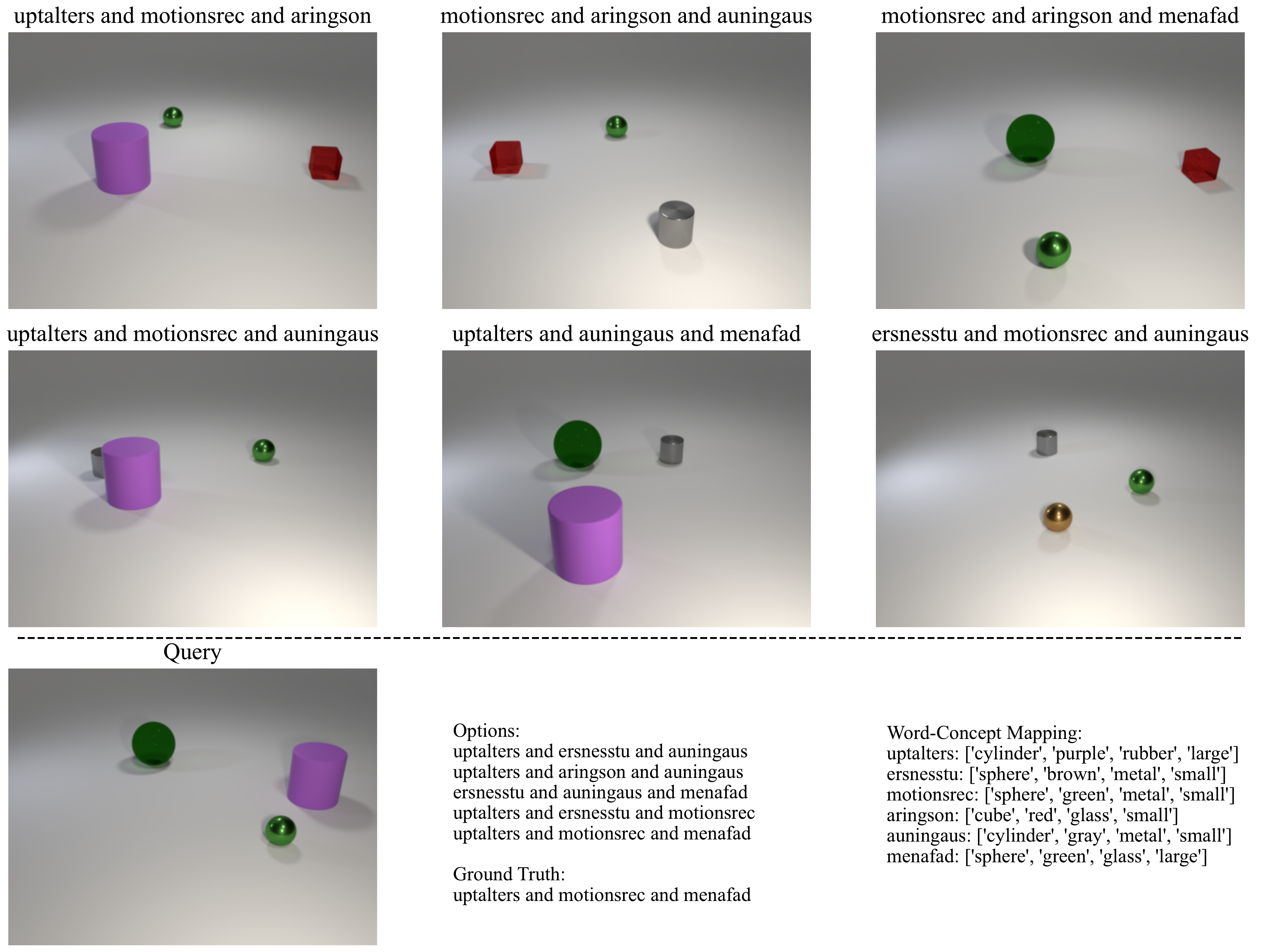}
    \includegraphics[width=0.83\linewidth]{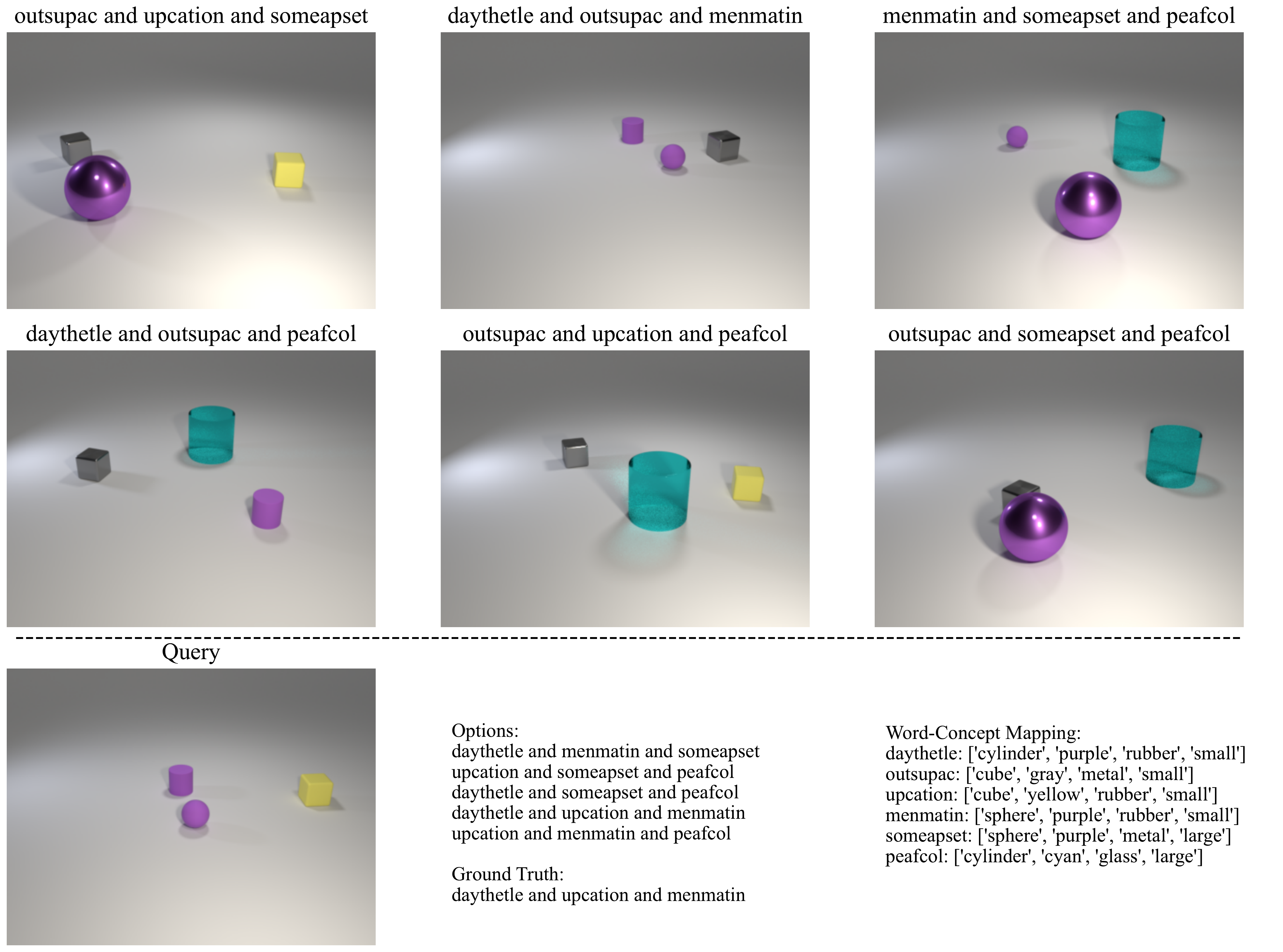}
\end{figure}

\subsection{\tcomposite{}}
\begin{figure}[ht!]
    \centering
    \includegraphics[width=0.83\linewidth]{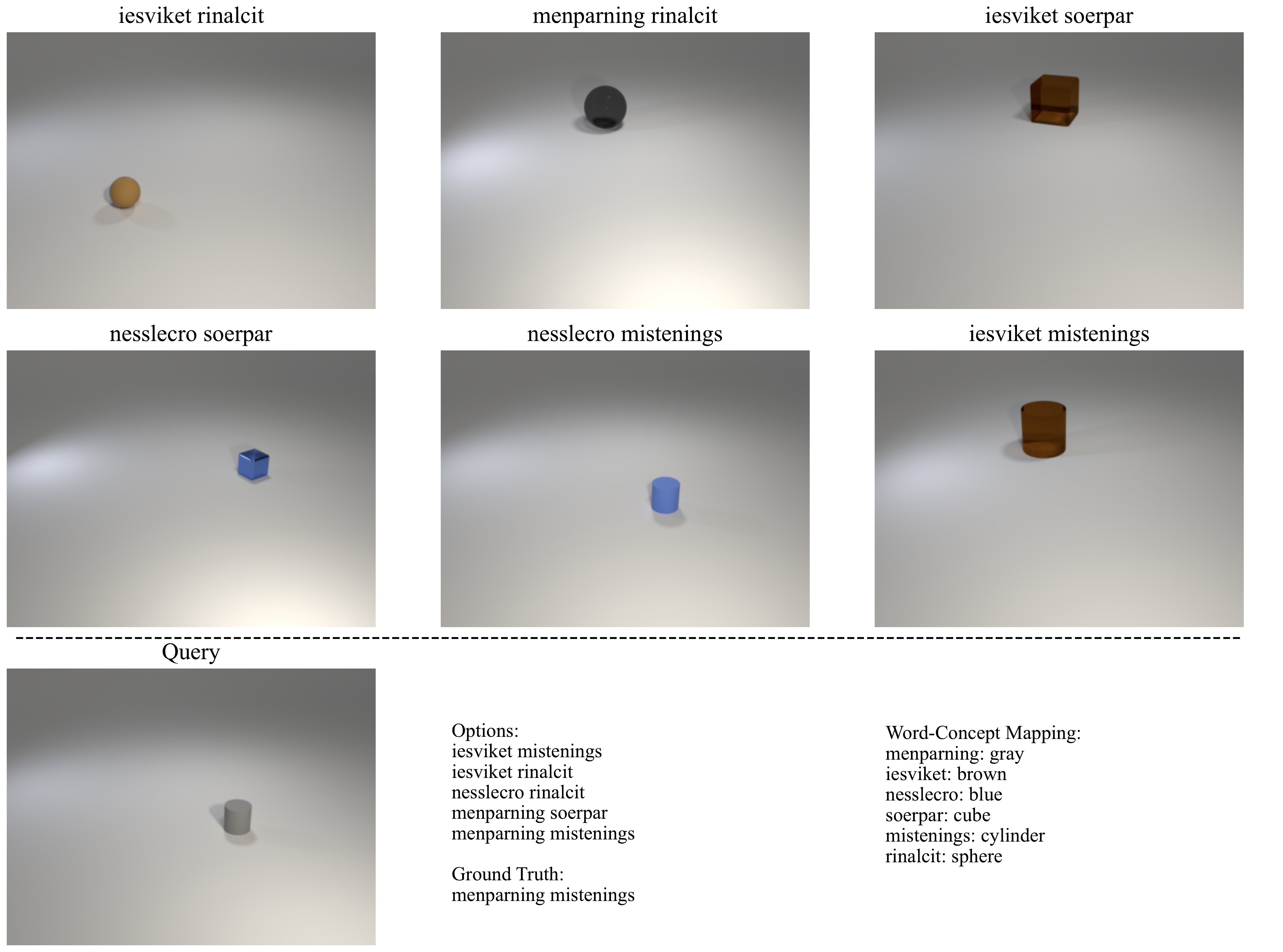}
    \includegraphics[width=0.83\linewidth]{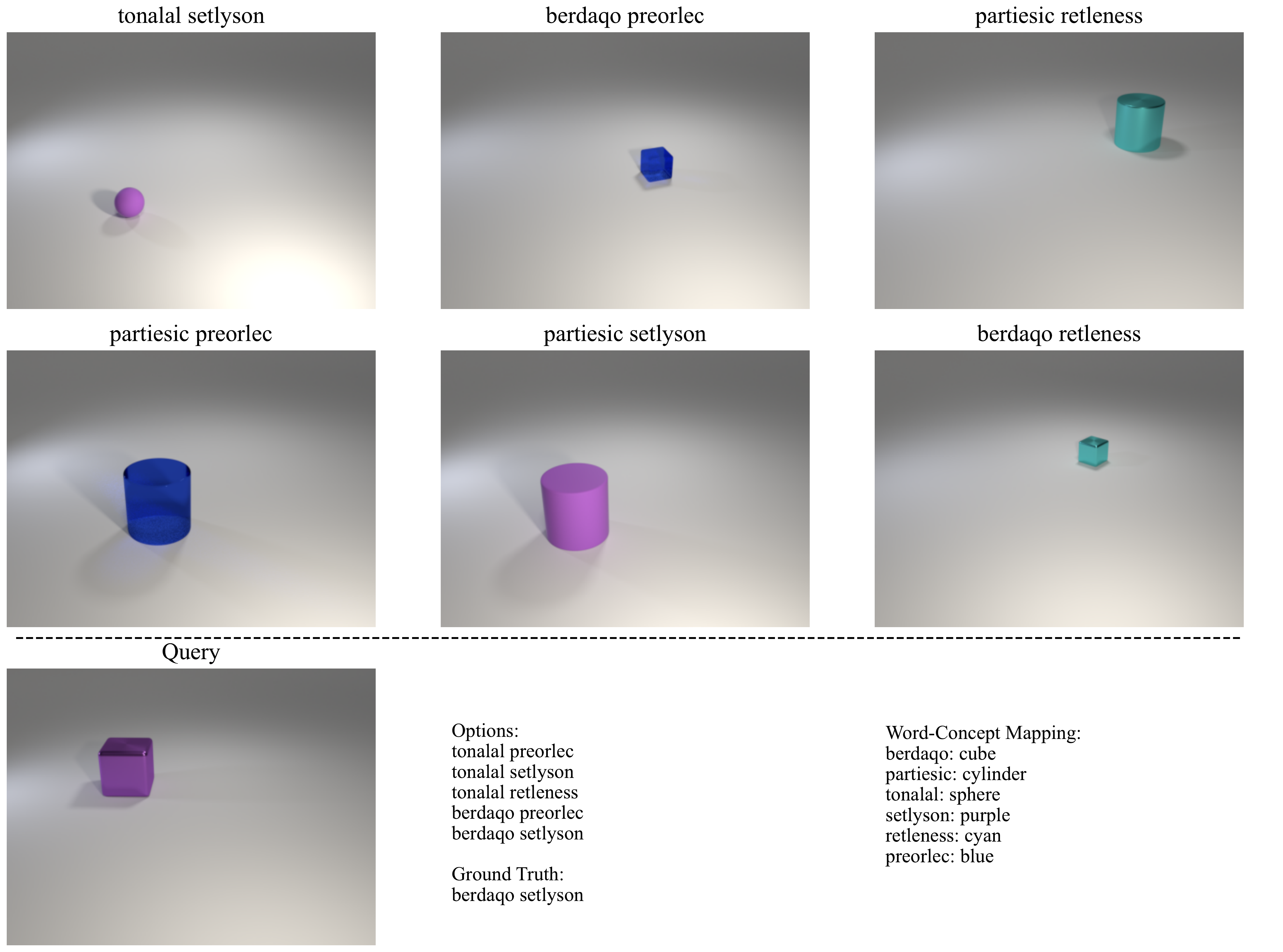}
\end{figure}

\subsection{\trelation{}}
\begin{figure}[ht!]
    \centering
    \includegraphics[width=0.83\linewidth]{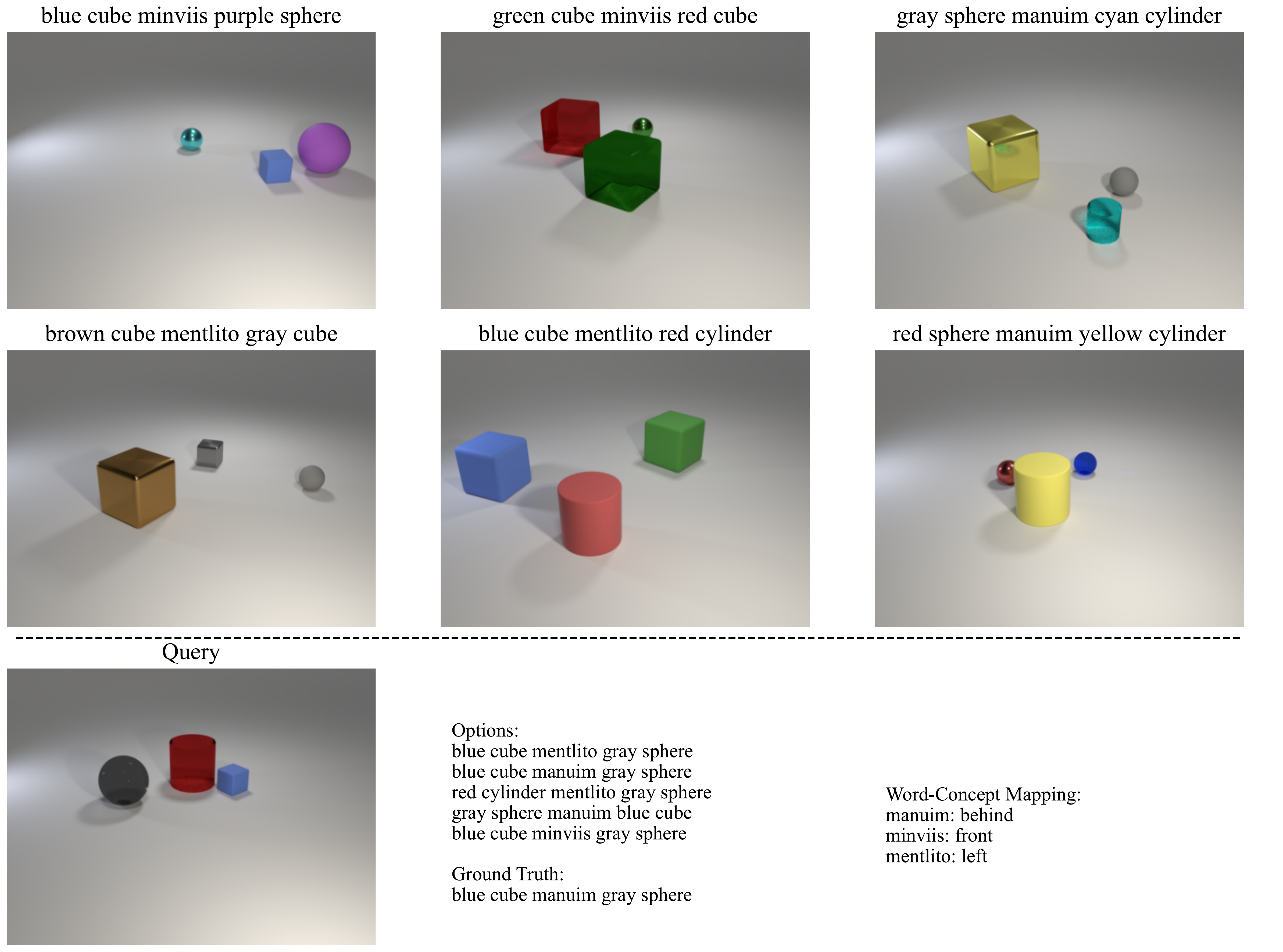}
    \includegraphics[width=0.83\linewidth]{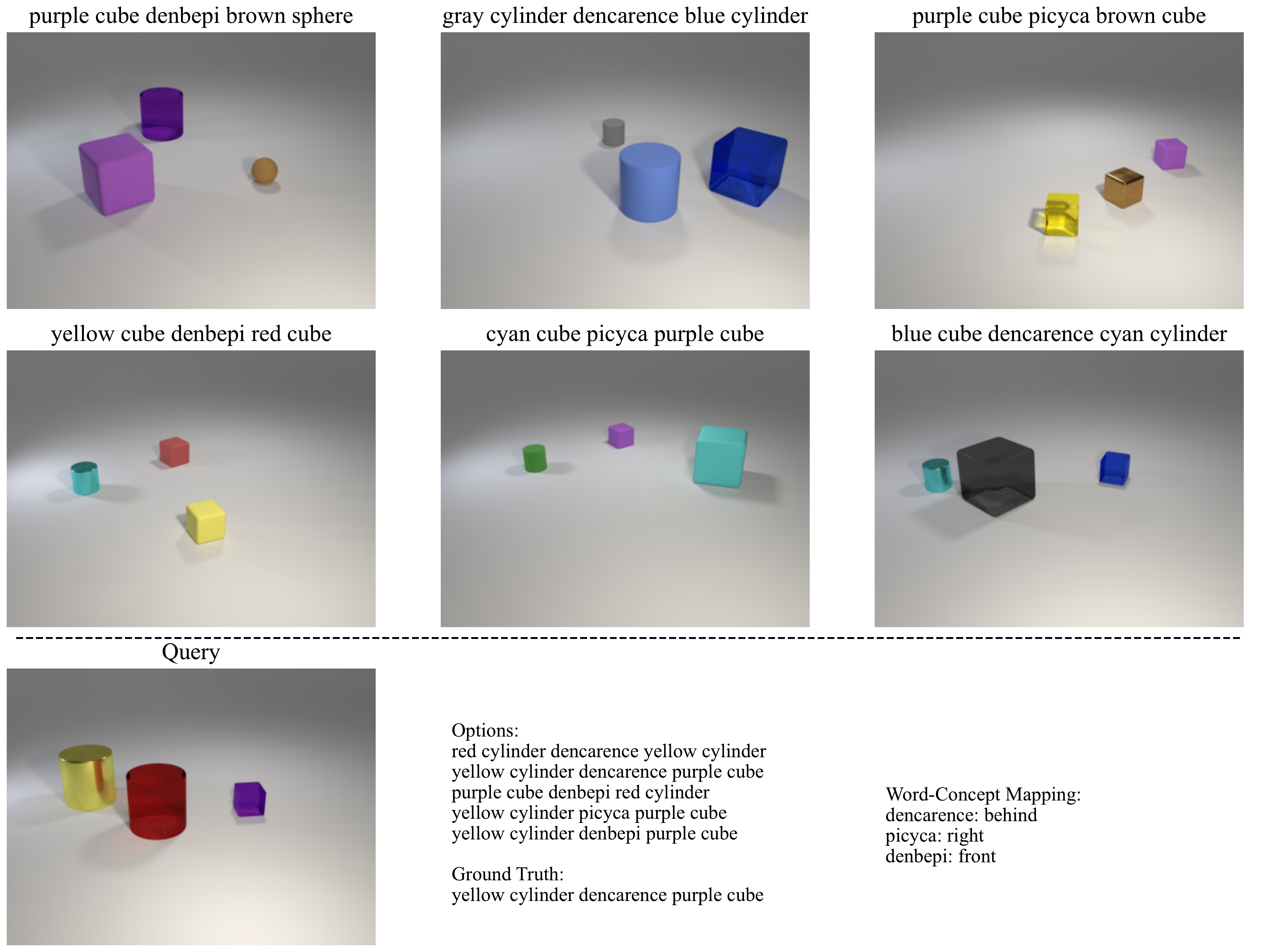}
\end{figure}
\clearpage

\subsection{\tbootstrap{}}
\begin{figure}[ht!]
    \centering
    \includegraphics[width=0.83\linewidth]{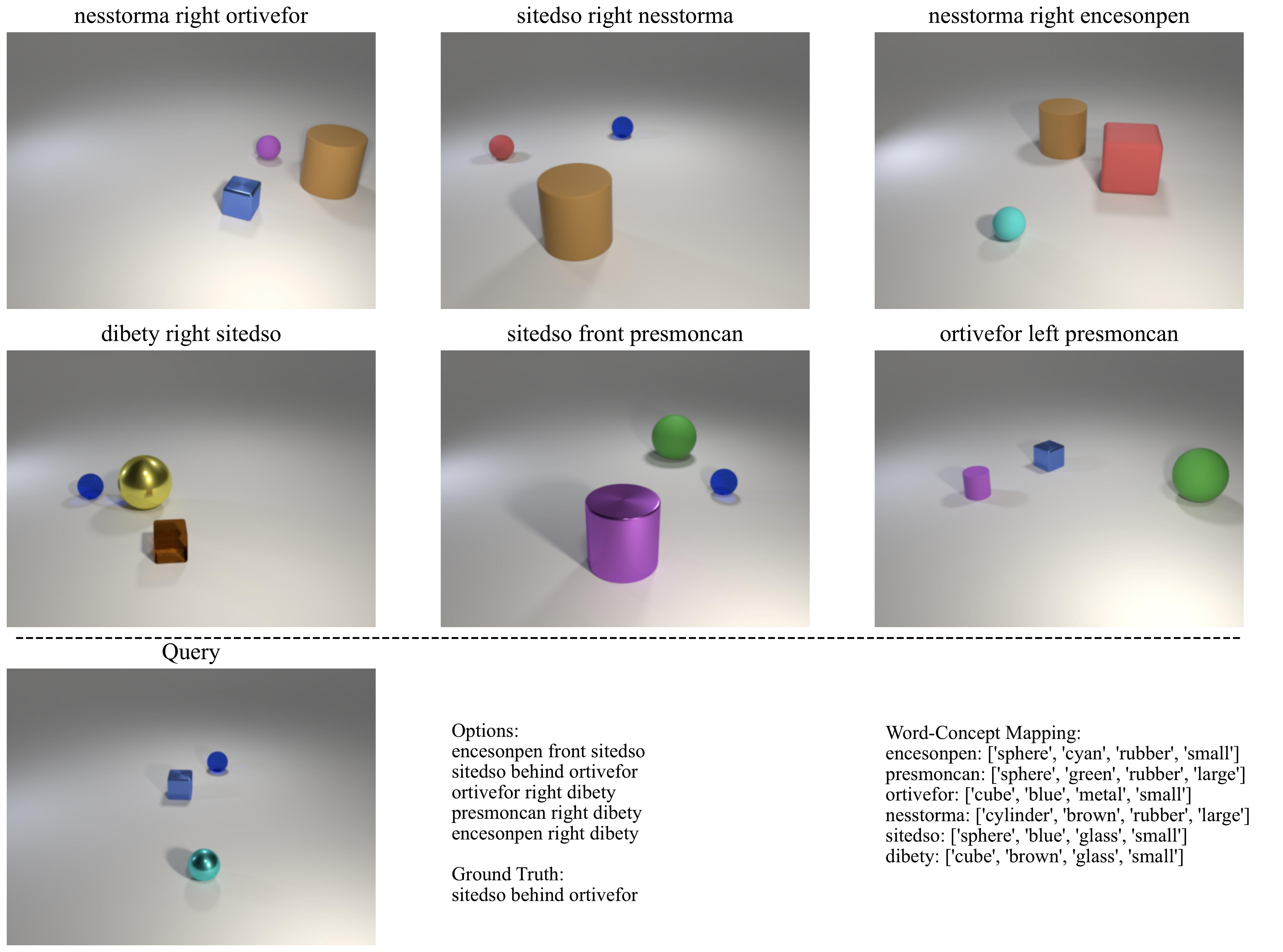}
    \includegraphics[width=0.83\linewidth]{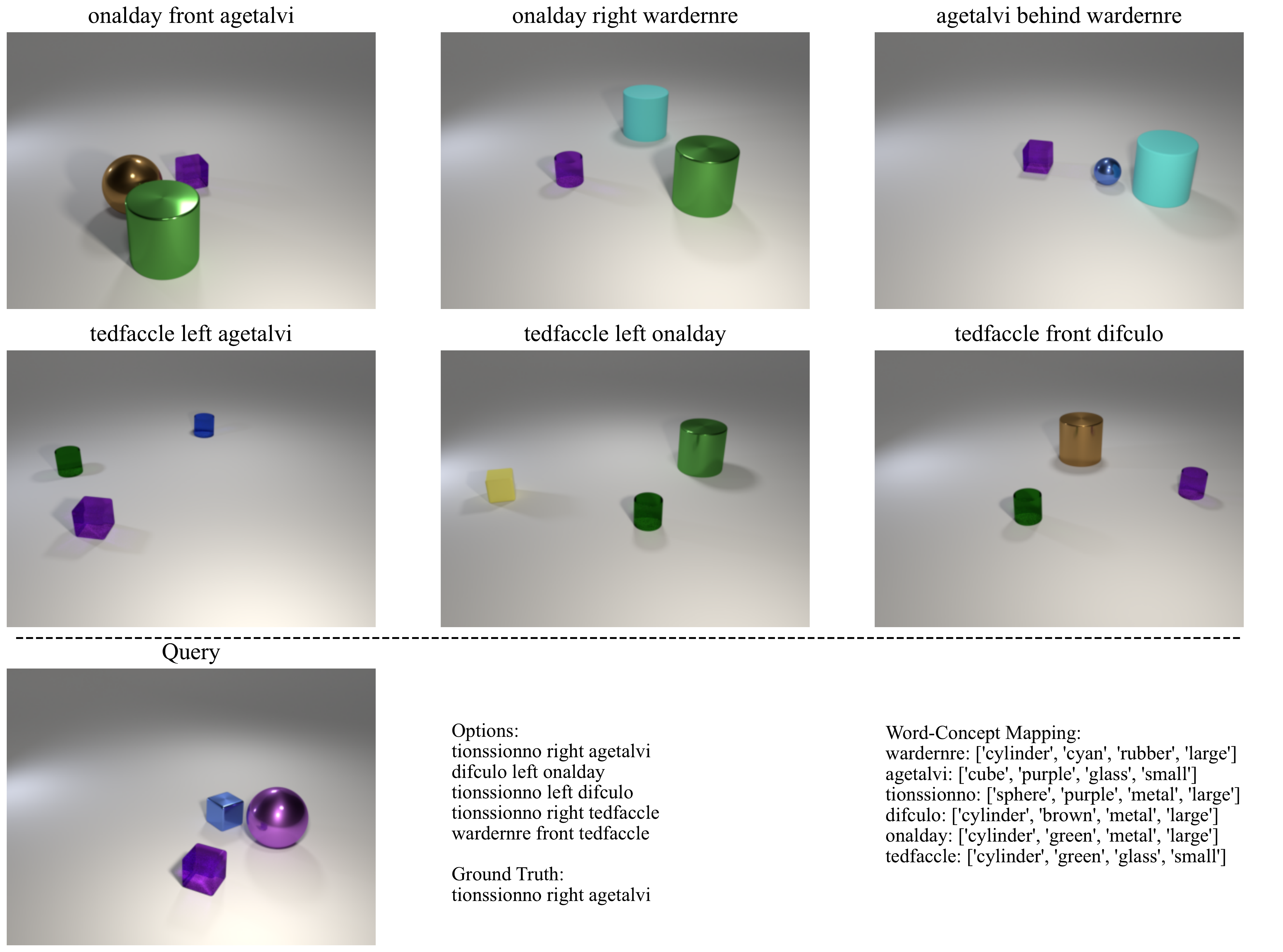}
\end{figure}

\subsection{\tnumber{}}
\begin{figure}[ht!]
    \centering
    \includegraphics[width=0.83\linewidth]{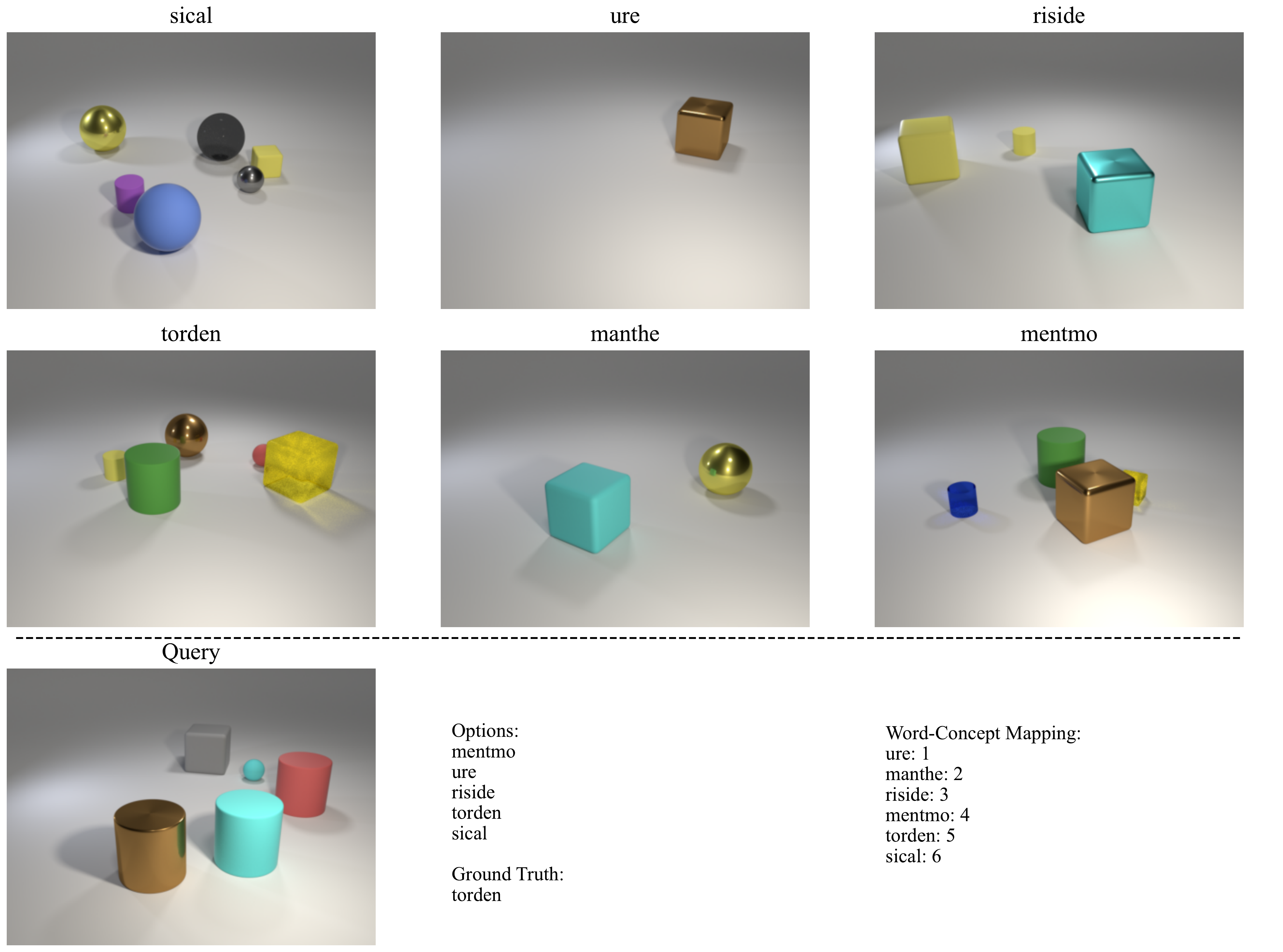}
    \includegraphics[width=0.83\linewidth]{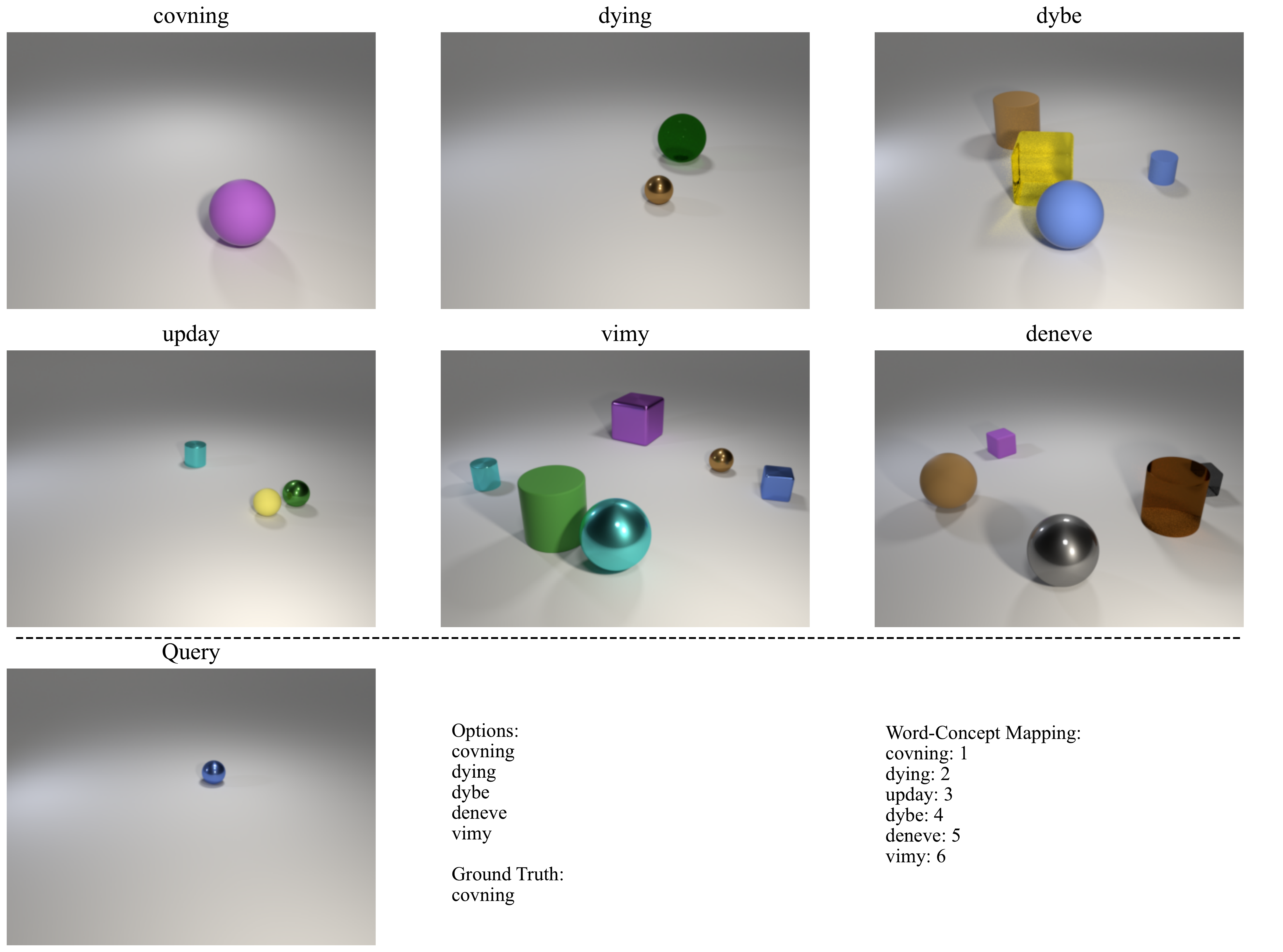}
\end{figure}

\subsection{\tpragmatic{}}
\begin{figure}[ht!]
    \centering
    \includegraphics[width=0.83\linewidth]{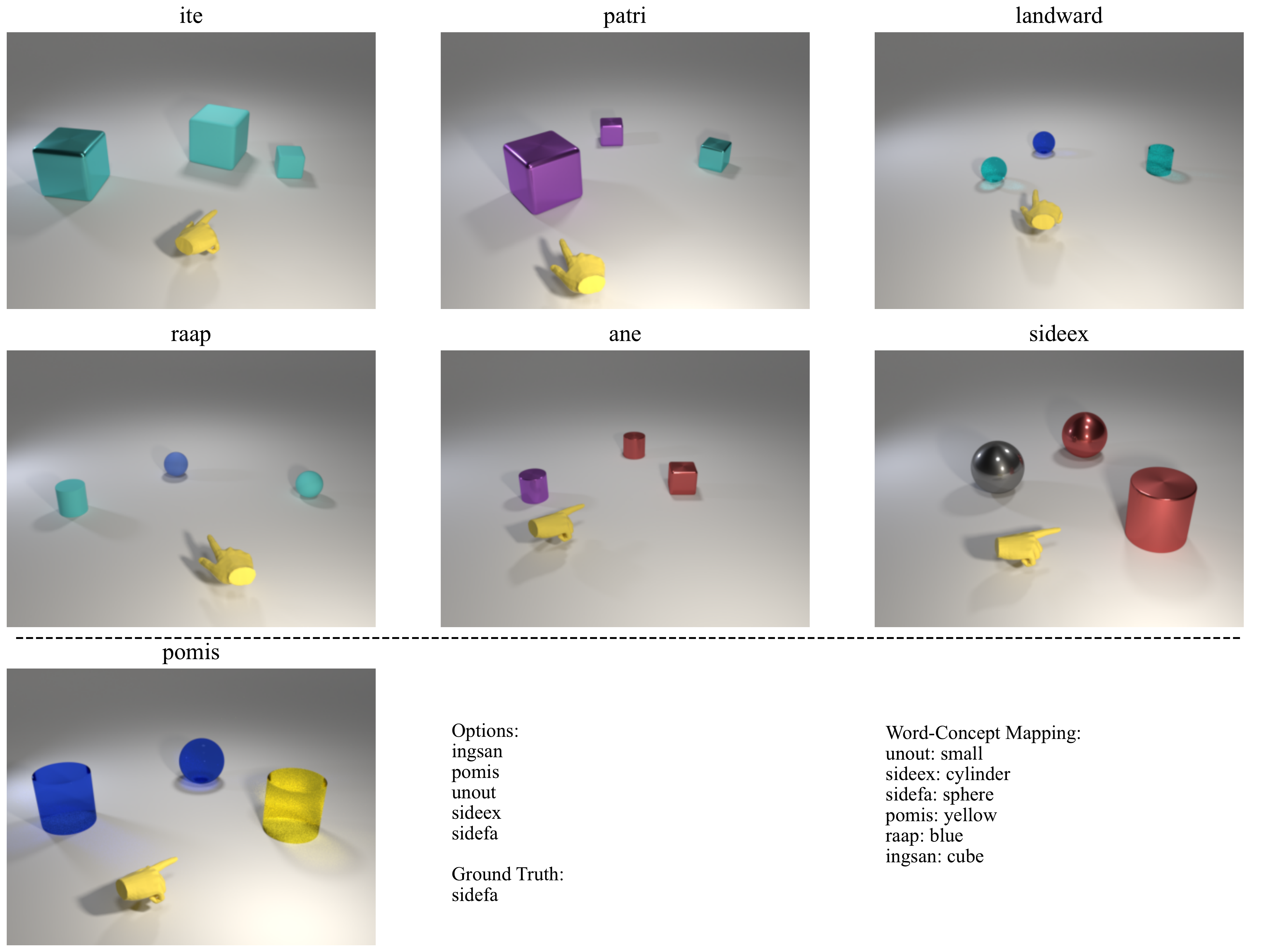}
    \includegraphics[width=0.83\linewidth]{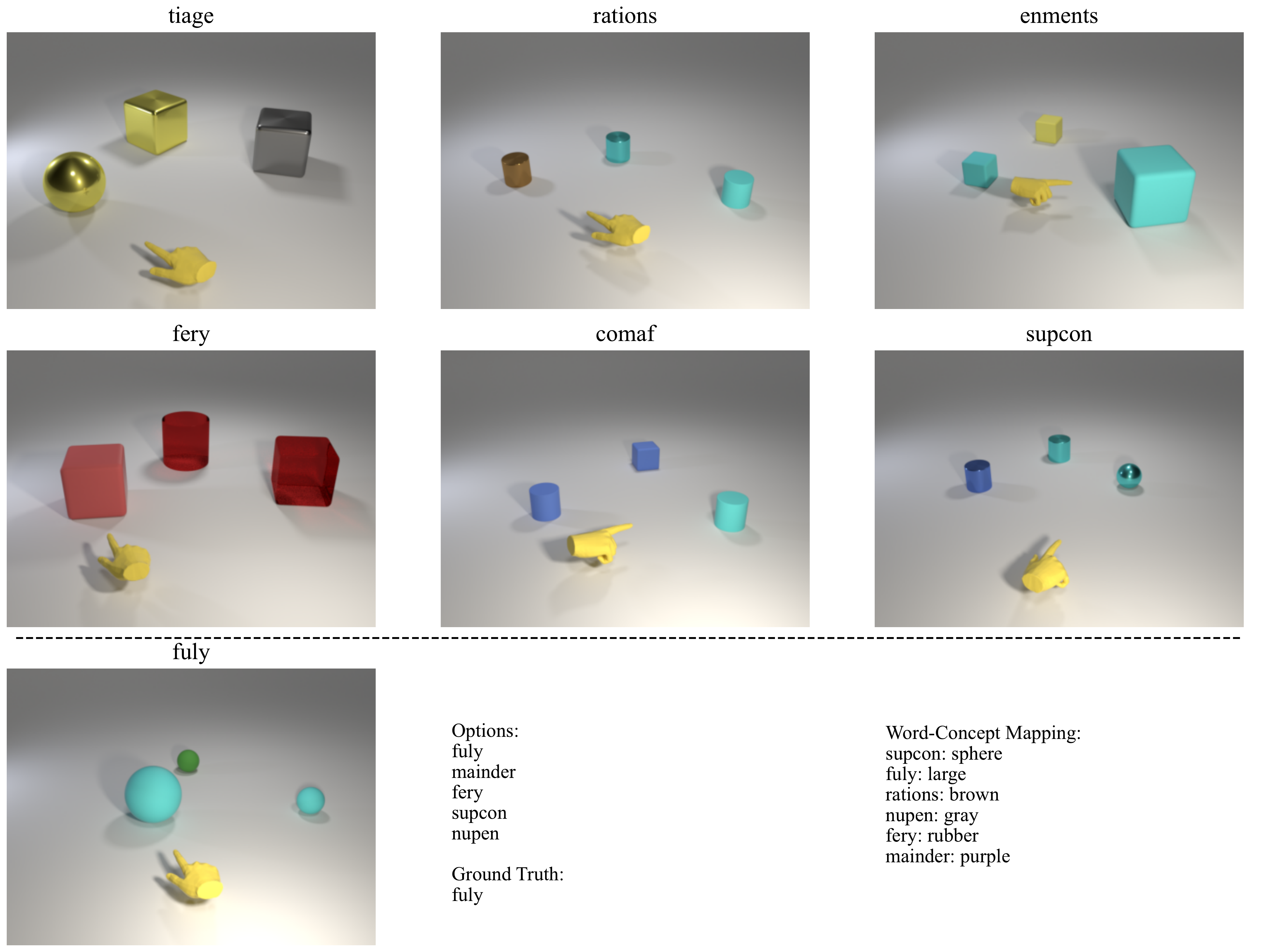}
\end{figure}